\documentclass[10pt,journal,compsoc]{IEEEtran}
\usepackage[utf8]{inputenc} 
\usepackage[T1]{fontenc}  

\usepackage[breaklinks=true,
            colorlinks,
            linkcolor = red,
            urlcolor  = purple, 
            citecolor = blue,
            bookmarks = false]{hyperref}  
\usepackage{url}  
\usepackage{amssymb}   
\usepackage{booktabs} 
\let\appendices\relax  
\usepackage{amsfonts}  
\usepackage{nicefrac}  
\usepackage{microtype}  
\usepackage{wrapfig}
\usepackage{bbding}
\usepackage{appendix}
\usepackage{graphicx}
\usepackage{amsmath}
\usepackage{amsthm}
\usepackage{subcaption}
\usepackage{float}
\usepackage{appendix}
\usepackage{xcolor}
\usepackage{multirow}

\usepackage{bm}
\usepackage{bbm}
\usepackage{makecell}
\usepackage{mathtools}
\usepackage{algorithmic}
\usepackage{algorithm}
\usepackage[algo2e,linesnumbered]{algorithm2e}

\newtheorem{lemma}{Lemma}

\newtheorem{definition}{Definition}
\newtheorem{corollary}{Corollary}
\newtheorem{remark}{Remark}

\SetArgSty{textrm}
\ifCLASSOPTIONcompsoc
  \usepackage[nocompress]{cite}
\else
  \usepackage{cite}
\fi

\ifCLASSINFOpdf
\else
\fi

\begin{document}

%
\title{Regularly Truncated M-estimators \\for Learning with Noisy Labels}
%
%
%
%

\author{Xiaobo~Xia\textsuperscript{$\ast$}\thanks{$\ast$\quad \ Equal contributions.},
   Pengqian~Lu\textsuperscript{$\ast$},
   Chen~Gong,~\IEEEmembership{Senior Member, IEEE,}
   Bo~Han,\\
  Jun~Yu,~\IEEEmembership{Senior Member, IEEE,}
  Jun~Yu$^\dagger$,
  Tongliang Liu,~\IEEEmembership{Senior Member, IEEE}
\thanks{$^\dagger$\quad \ Corresponding author.}
\IEEEcompsocitemizethanks{
\IEEEcompsocthanksitem X. Xia and T. Liu are with the Sydney AI Center, School of Computer Science, Faculty of Engineering, The University of Sydney, Darlington, NSW2008, Australia (e-mail: xxia5420@uni.sydney.edu.au;  tongliang.liu@sydney.edu.au).
\IEEEcompsocthanksitem P. Lu is with the Australian AI Institute, Faculty of Engineering and IT, The University of Technology Sydney, Broadway, NSW, 2007, Australia (e-mail: pengqian.lu@student.uts.edu.au).
\IEEEcompsocthanksitem C. Gong is with the PCA Lab, the Key Laboratory of Intelligent Perception and Systems for High-Dimensional Information of Ministry of Education, School of Computer Science and Engineering, Nanjing University of Science and Technology, Nanjing, 210094, P.R. China; and is also with the Department of Computing, Hong Kong Polytechnic University, Hong Kong SAR, China (e-mail:chen.gong@njust.edu.cn).
\IEEEcompsocthanksitem B. Han is with the Department of Computer Science, Hong Kong Baptist University, Hong Kong, China (email: bhanml@comp.hkbu.edu.hk).
\IEEEcompsocthanksitem J. Yu is with the School of Computer Science and Technology, Hangzhou Dianzi University, Hangzhou, 310018, China (e-mail: yujun@hdu.edu.cn).
\IEEEcompsocthanksitem J. Yu is with the Department of Automation, University of Science and Technology of China, Hefei, 230026, China (e-mail: harryjun@ustc.edu.cn).}

}

\markboth{Regularly Truncated M-estimators
for Learning with Noisy Labels}%
{Regularly Truncated M-estimators
for Learning with Noisy Labels}

\IEEEtitleabstractindextext{%
\begin{abstract}
 The \textit{sample selection} approach is very popular in learning with noisy labels. As deep networks \textit{``learn pattern first''}, prior methods built on sample selection share a similar training procedure: the small-loss examples can be regarded as clean examples and used for helping generalization, while the large-loss examples are treated as mislabeled ones and excluded from network parameter updates. However, such a procedure is \textit{arguably debatable} from two folds: (a) it does not consider the bad influence of noisy labels in selected small-loss examples; (b) it does not make good use of the discarded large-loss examples, which may be clean or have meaningful information for generalization. In this paper, we propose regularly truncated M-estimators (RTME) to address the above two issues \textit{simultaneously}. Specifically, RTME can \textit{alternately switch modes between truncated M-estimators and original M-estimators}. The former can \textit{adaptively} select small-losses examples without knowing the noise rate and reduce the side-effects of noisy labels in them. The latter makes the possibly clean examples but with large losses involved to help generalization. Theoretically, we demonstrate that our strategies are label-noise-tolerant. Empirically, comprehensive experimental results show that our method can outperform multiple baselines and is robust to broad noise types and levels. The
implementation is available at \href{https://github.com/xiaoboxia/RTM_LNL}{https://github.com/xiaoboxia/RTM\_LNL}.
\end{abstract}
}

\maketitle

\IEEEdisplaynontitleabstractindextext

\IEEEpeerreviewmaketitle

\IEEEraisesectionheading{\section{Introduction}\label{sec:introduction}}
\IEEEPARstart{L}{earning} with noisy labels is one of the hottest problems in weakly supervised learning \cite{zhou2018brief,yan2023mutual,paleka2023law,olmin2022robustness,jiang2020beyond}, since noisy labels are ubiquitous in real-world datasets, which always arise in mistakes of manual or automatic annotators \cite{he2010maximum,he2013half,ke2020laplacian,patel2023adaptive,iscen2022learning,bae2022noisy,liang2022few,yang2023parametrical,silva2022noise}. Noisy labels can impair the performance of models, especially deep learning models (e.g., convolutional and recurrent neural networks) which have large model capacities. General regularization techniques such as dropout and weight decay cannot address this issue well \cite{zhang2017understanding}. Different approaches therefore have been proposed for robust learning with noisy labels \cite{li2021provably,xia2020part,bai2021me,wu2018light,xie2022ccmn,xie2021partial,dai2022towards,wang2023promix,bucarelli2023leveraging}. Among them, the \textit{sample selection} approach attracted a lot of attention from researchers, since it always has a simple mechanism but promising performance, and is orthogonal to other approaches \cite{yao2020searching,yu2019does,feng2023ot}. This approach is also \textit{our focus} in this paper. 

\begin{figure}[!h]
\centering
\begin{subfigure}{.42\linewidth}\label{fig:motivation_a}
  \centering
  \includegraphics[width=1\linewidth]{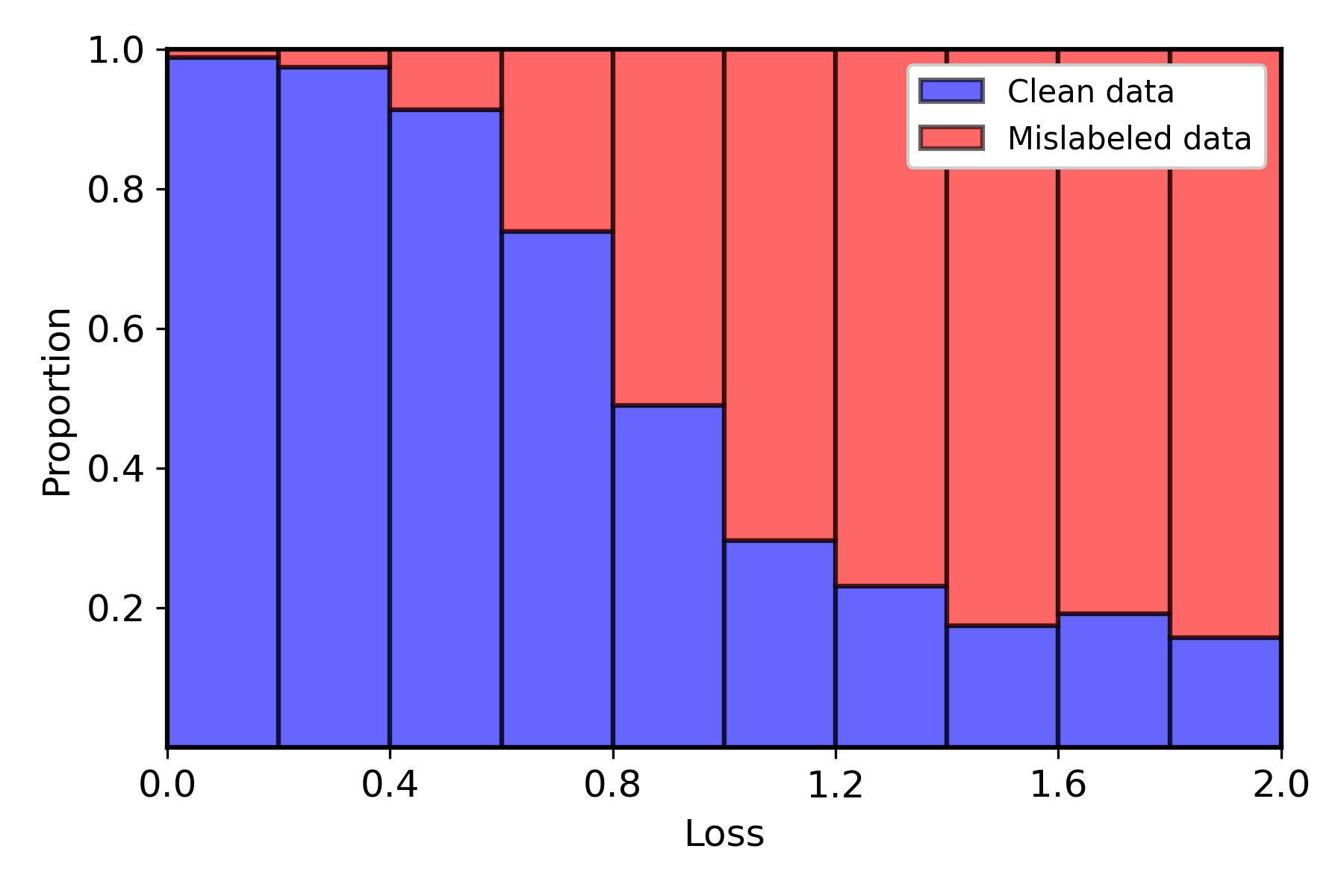} 
  \caption{}
\end{subfigure}
\hspace{4mm}
\begin{subfigure}{.42\linewidth}\label{fig:motivation_b}
  \centering
  \includegraphics[width=1\linewidth]{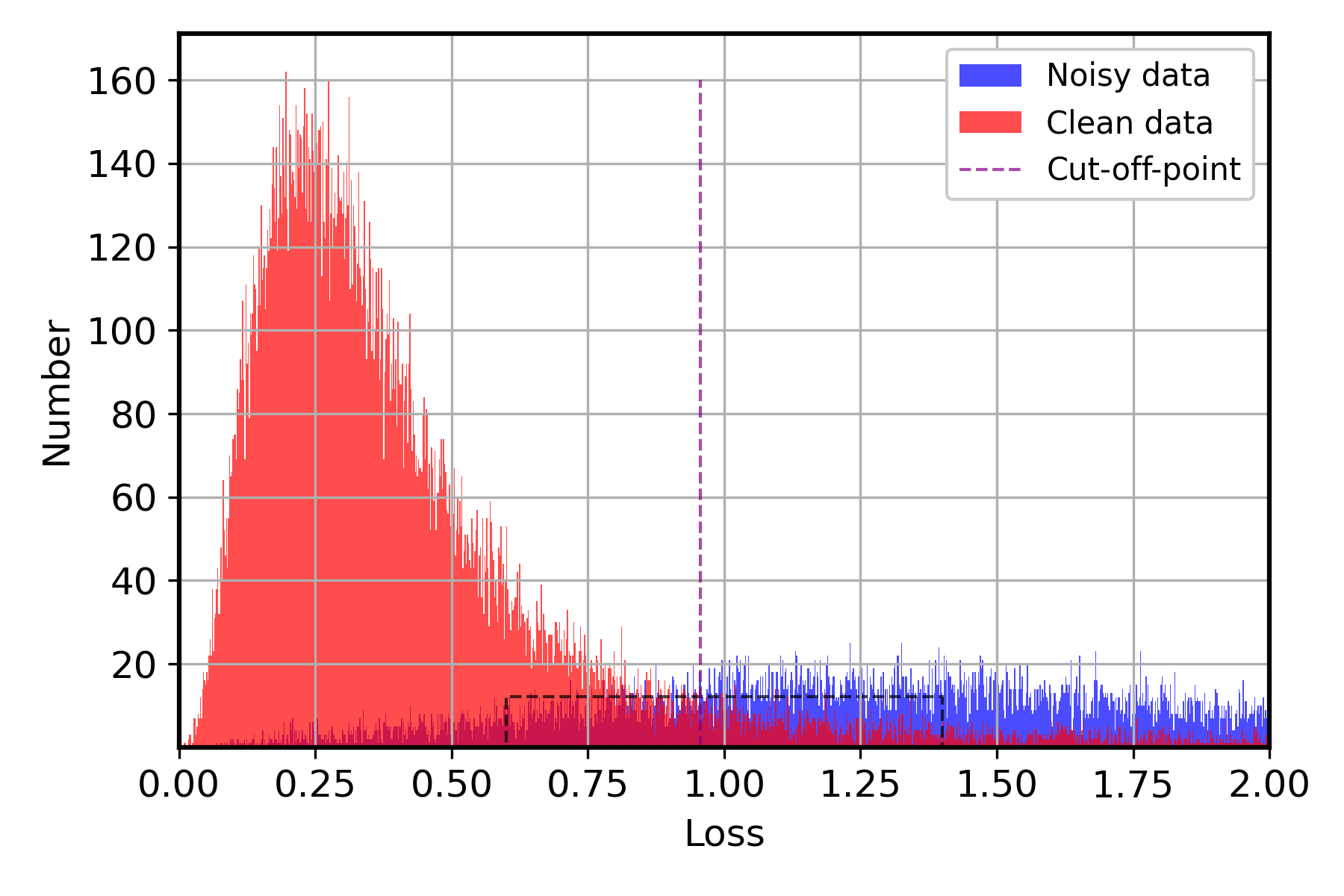} 
  \caption{}
\end{subfigure}
\vspace{-5pt}
\caption{\small{Illustrations of training loss distributions. Experiments were conducted on synthetic \textit{CIFAR-10} with instance-dependent label noise \cite{xia2020part}. The noise rate is set to 30\%. \textbf{(a):} Loss \textit{vs} Proportion of clean/mislabeled data. Here, \textit{proportion=(\# clean (resp. mislabeled data)) / (\# all training data)}. The proportion of clean data is almost \textit{negatively related} to the values of losses. \textbf{(b):} Loss \textit{vs} Number. Noisy labels still exist in the selected small-loss examples, which hurt generalization.}}
\vspace{-7pt}
\label{fig:motivation}
\end{figure}

The sample selection approach is based on selecting possibly clean examples from noisy examples for training.  Intuitively, if we can exploit less noisy data for network parameter updates, the network will be more robust. At the present stage, the sample selection built on the small-loss criteria is the most common method, and has been verified to be effective in many circumstances \cite{jiang2018mentornet,han2018co,yu2019does,xia2021instance,wei2020combating}.  Specifically, since deep networks \textit{learn patterns first} \cite{zhang2017understanding}, they would first memorize training data of clean
labels and then those of noisy labels with the assumption that clean labels are of the majority in a noisy class. Small-loss examples can be regarded as clean examples \textit{with high probability}. Therefore, in each iteration for a mini-batch data, the small-loss examples are selected for robust training \textit{with equal importance}. By contrast, the large-loss examples are \textit{treated to be mislabeled} and \textit{excluded from training}.

However, such a selection procedure is \textit{debatable} from \textit{two folds}. First, the equal importance should not be assigned to different small-loss examples. Specifically, although we \textit{rank} the losses of all examples and regard a proportion of examples as clean examples, such a way does not guarantee that selected examples are \textit{completely clean} \cite{han2018co}, especially the examples have \textit{relatively large} losses but still are seen to be clean \cite{arazo2019unsupervised}. Recall the selection procedure, the purity of an example is \textit{negatively correlated} with its loss, i.e., the example with a smaller loss is regarded to be clean with a higher degree of confidence (Fig.~\ref{fig:motivation}). Therefore, we should assign larger weights to the examples with smaller losses to make use of these ``confirmable'' clean examples to help generalization. Second, the large-loss examples should not be discarded directly. Specifically, although the large-loss examples may be mislabeled, the instances (e.g., images) may be helpful for generalization \cite{han2020sigua}. This opinion is motivated by the prior work \cite{xia2021instance}, which shows that the images of mislabeled data may have meaningful information (e.g., scene information), even though such images come from a \textit{different instance space}. For our task, mislabeled examples and clean examples share the same instance space. Such mislabeled examples is thereby more reasonable to be considered useful, and can be exploited for training. 

In this paper, to relieve the above two issues simultaneously while keeping \textit{end-to-end}, we propose \textit{regularly truncated M-estimators}. More specifically, we borrow the statistical robust M-estimators in statistical learning \cite{zhang2014novel}, which can adaptively assign larger weights to examples with smaller losses. Based on the multiple robust M-estimators, to perform sample selection, we develop novel truncated M-estimators. By performing truncation on magnitudes of losses meanwhile \textit{without knowing or estimating the noise rate}, our truncated M-estimators can concern the purity of small-loss examples and assign \textit{zero weights} to possibly mislabeled examples to enhance networks. Since truncated M-estimators only consider the better use of small-loss examples, but do not make use of meaningful large-loss examples, we \textit{regularly} switch robust M-estimators between truncated ones and original ones to achieve the proposed regularly truncated M-estimators. In this way, we can assign different weights to ``clean'' examples after sample selection (with truncated M-estimators). Additionally, the large-loss examples can be introduced regularly into network parameter updates for helping generalization (with original robust M-estimators). As large-loss examples are not introduced into training all the time, but are introduced regularly, and have \textit{smaller} weights compared with small-loss examples, the side effect of possibly mislabeled examples can be reduced effectively, following better generalization. 

Before delving into details, we highlight the main contributions of this paper in three folds:
\begin{itemize}
    \item  We show that the most frequently used sample selection procedure still has some potential weaknesses and discuss them carefully. Based on this, novel regularly truncated M-estimators are proposed to address the mentioned issues.
    \item Theoretical analysis is presented to demonstrate that the proposed methods are label-noise-tolerant. We also discuss that this work actually provides a new and interesting perspective to make one loss function robust to label noise using the truncation mechanism.
    \item Extensive experiments on datasets with synthetic label noise and real-world label noise are conducted to verify the effectiveness of the proposed methods. Experimental results justify our claims well. Codes are open-source for future research. 
\end{itemize}

\subsection{Previous work}
In this subsection, we briefly review prior approaches to learning with noisy labels, including robust loss functions, loss correction, and label correction. Our focus, i.e., the sample selection approach, will be introduced in detail later.

\noindent\textbf{Robust loss functions.}  Some efforts have been made to design robust loss functions to handle noisy labels, e.g., the generalized cross-entropy loss \cite{zhang2018generalized}, the normalized loss \cite{ma2020normalized}, the curriculum loss \cite{lyu2019curriculum}, the symmetric (cross-entropy) loss \cite{wang2019symmetric,charoenphakdee2019symmetric}, the negative loss \cite{kim2019nlnl}, the peer loss \cite{liu2019peer}, and the mutual information loss \cite{xu2019l_dmi}, etc. 

\noindent\textbf{Loss correction.} This approach improves the robustness of networks by modifying the training loss. The modification can be achieved by reweighting losses \cite{liu2016classification,ren2018learning}, estimating the noise transition matrix \cite{xia2020part,wu2020class2simi,xia2022extended,yao2020dual,zhu2021clusterability,zhu2021second,liu2022identifiability,kye2022learning}, and adding an adaption layer \cite{goldberger2016training}, etc. 

\noindent\textbf{Label correction.} The label correction approach~\cite{yi2019probabilistic,zhang2021learningwith} aims to correct wrong labels into correct ones. The correction can be obtained by using directed graphical models \cite{xiao2015learning}, conditional random fields \cite{vahdat2017toward},  knowledge graphs \cite{li2017learning}, and joint optimization methods \cite{tanaka2018joint}, etc. 

\noindent\textbf{Integrated approach.} Nowadays, state-of-the-art methods of handling noisy labels~\cite{li2020dividemix,li2022selective,huang2023twin} are often designed by integrating various techniques at the same time. For example, they can simultaneously involve Mixup \cite{zhang2017mixup}, soft labels \cite{reed2014training}, and semi-supervised learning \cite{berthelot2019mixmatch}, or involve sample selection and self-supervised learning \cite{wang2023mosaic}. We suggest that the readers refer to \cite{han2020survey,song2022learning} for more details about learning with noisy labels.

Compared with these prior effects, this paper offers an inspiring perspective to handle noisy labels, i.e., regularly truncated M-estimators, which successfully connects the classical statistical M-estimators and learning with noisy labels. Conceptually, this connection is new and valuable, and contributes to the research field.

\subsection{Organization}
 The rest of this paper is organized as follows. In Section \ref{sec:2}, we introduce the problem setting and some background of the proposed methods. In Section \ref{sec:3}, we present the proposed methods formally. Experimental results are discussed in Section \ref{sec:4}. The conclusion is given in Section \ref{sec:5}. 
\section{Preliminaries}\label{sec:2}
In this section, we first introduce the notations (Section \ref{sec:2.1}) and problem setting (Section \ref{sec:2.2}). Then the sample selection approach for learning with noisy labels is discussed in detail (Section \ref{sec:2.3}). Finally, we provide a brief introduction for the M-estimator (Section \ref{sec:2.4}) and employed M-estimators in this work (Section \ref{sec:2.5}). 
\subsection{Notations}\label{sec:2.1}
Vectors and matrices are denoted by bold-faced letters. We use $\|\cdot\|_p$ as the $\ell_p$ norm
of vectors or matrices. Let $[z]=\{1,2,\ldots,z\}$. For a function $g$, we use $\nabla g$ to denote its gradient. Let $\mathbbm{1}_{\{\cdot\}}$ be the indicator function and ``\text{mod}'' be the math operation of taking the remainder.

\subsection{Problem setup}\label{sec:2.2}
Let $\mathcal{X}$ and $\mathcal{Y}$ be the instance and label space respectively. We consider a $k$-class classification problem, i.e., $\mathcal{Y}=[k]$.  Let $(\bm{x},y)$ be the random variable pair of interest, and $p(x,y)$ be the
underlying joint density from which test data will be sampled. In \textit{learning with noisy labels}, the labels of training data are corrupted. The training data are sampled from a \textit{corrupted joint density} $p(\bm{x},\tilde{y})$ rather than $p(\bm{x},y)$, where $\tilde{y}$ denotes the random variable of the noisy labels. Here, $p(\bm{x})$ remains the same, but $p(y|\bm{x})$  is corrupted into $p(\tilde{y}|\bm{x})$~\cite{patrini2017making,han2020sigua}. Therefore, we have an observed noisy training sample as follows:
\begin{equation}
    S=\{(\bm{x}_i,\tilde{y}_i)\}_{i=1}^n\stackrel{\rm i.i.d.}{\sim}p(\bm{x},\tilde{y})=p(\tilde{y}|\bm{x})p(\bm{x}),
\end{equation}
where $n$ denotes the sample size of training data. 

Let $f:\mathcal{X}\rightarrow\mathbbm{R}^k$ be a classifier with parameters $\bm{w}$. Let $\ell:\mathbbm{R}^k\rightarrow\mathbbm{R}$ be a surrogate loss function for $k$-class classification. In this paper, we use the \textit{softmax cross entropy loss} (abbreviated as the CE loss) \cite{mohri2018foundations}. Given an arbitrary training example $(\bm{x}_i,\tilde{y}_i)$, with parameters $\bm{w}$, we can obtain its CE loss:
\begin{equation}
    L_i=\ell(f(\bm{w};\bm{x}_i),\tilde{y}_i). 
\end{equation}

\subsection{Sample selection for handling noisy labels}\label{sec:2.3}
Prior effects exploited the sample selection approach to handle noisy labels \cite{jiang2018mentornet,han2018co,han2020sigua,yu2019does,xia2021sample,wang2018iterative}, which only used the ``clean” examples (with relatively small losses) from each mini-batch for training. These clean examples have the same weights to contribute to optimization. Such methods employ the memorization effects of deep networks \cite{zhang2017understanding}, which show that they would first memorize training data with clean labels and then those with noisy labels. We use a self-teach version of MentorNet \cite{jiang2018mentornet} to give a better understanding for readers. The main procedure is shown in Algorithm \ref{alg:gen}.

Let us look at this procedure more closely. When a mini-batch data are formed (Step 5), we start to select possibly clean examples. In \textbf{Step 6}, we select a proportion of small-loss examples (controlled by the function $R(T)$) based on the network predictions. The large-loss examples are \textit{abandoned directly} from optimization. In \textbf{Step 7}, the selected small-losses examples in the previous step are exploited for parameter updates. Their importance is seen to be the \textit{same} for generalization. In Step 9, we update $R(T)$. Note that the function $R(T)$ needs to be designed carefully to better use the memorization effects of deep networks, and always is task-dependent \cite{yao2020searching}. For instance, in \cite{han2018co,yu2019does,wei2020combating}, $R(T)=1-\min\{T/T_k*\tau,\tau\}$, where $\tau$ is the noise rate. In practice, we cannot know the noise rate and have to estimate it \cite{arazo2019unsupervised}. Unfortunately, in some cases, e.g., the label noise is \textit{instance-dependent}, the noise rate is hard to be estimated accurately \cite{xia2020part,cheng2017learning}. Accordingly, the effect of the sample selection process will be influenced, which is never our desideratum. 

As mentioned above, it is argued that the sample selection procedure (Algorithm~\ref{alg:gen}) does not take care of the mislabeled data in the selected one and does not make use of large-loss data. Our methods tackle the two issues directly and are more advanced in that (1) the side-effect of mislabeled data belonging to selected data is reduced; (2) the meaningful formation of large-loss examples can be employed to help generalization. The technical implementation of our methods will be carefully discussed later.

\begin{algorithm}[!t]
\caption{The main procedure of self-teach MentorNet for combating noisy labels.}
\begin{algorithmic}[1]
	    \STATE \textbf{Input}: initialized classifier $f$, epoch $T_k$ and $T_{\max}$, iteration $t_{\max}$.
		\FOR{$T = 0, \dots, T_{\max}-1$}
		\STATE \textbf{Shuffle} training dataset $S$;
		\FOR{$t = 0, \dots, t_{\max}-1$}
		\STATE \textbf{Draw} a mini-batch $\bar{S}$ from $S$;
		\STATE \textbf{Select} $R(T)$ small-loss examples $\bar{S}_f$ from $\bar{S}$ based on classifier's predictions;
		\STATE \textbf{Update} classifier parameters only using $\bar{S}_f$;
		\ENDFOR
		\STATE \textbf{Update} $R(T)$ with $T_k$;
		\ENDFOR
		\STATE \textbf{Output}: trained classifier $f$.
	\end{algorithmic}
	\label{alg:gen}
\end{algorithm}

\subsection{The M-estimator}\label{sec:2.4}
In statistics, M-estimators are a broad class of extremum estimators for which the objective function is a sample average~\cite{james2013introduction}. We use a classical example (i.e., the estimation of the geometric median) to give an explanation for the M-estimator (cf. \cite{zhang2014novel,he2013robust}). For a dataset $\mathcal{A}=\{\mathbf{a}_i\}_{i=1}^{N}\subset\mathbbm{R}^d$, the geometric median is the minimizer of the following function of $\mathbf{b}\in\mathbbm{R}^d$: 
\begin{equation}\label{eq:m_estimator}
    \sum_{i=1}^{N}\|\mathbf{b}-\mathbf{a}_i\|_2.
\end{equation}
This is a typical example of an M-estimator, that is a minimizer of a function of the form $\sum_{i=1}^N\rho(r_i)$, where $r_i$ is a residual of the $i$-th data point, from the parametrized object (\ref{eq:m_estimator}). We have $r_i=\|\mathbf{b}-\mathbf{a}_i\|_2$ and $\rho(r_i)=r_i$. If there are some outliers in $\mathcal{A}$, the residuals of some data points may be unusually large and cause the minimizer to be unable to be learned accurately. Therefore, we need to give smaller weights to such data points to make results more robust, i.e., using robust M-estimators. 

Below, we give a formal definition of the M-estimator in the context of learning with noisy labels. 
\begin{definition}[M-estimator]
In learning with noisy labels, an estimator is called the M-estimator, if it is an extremum
estimator and can improve the robustness of the model by mitigating the side effect of mislabeled data during empirical risk minimization.
\end{definition}

For our task, we use the CE loss to measure the difference between predictions and given labels, the minimizer is the classifier $f$. During empirical risk minimization, the loss of the $i$-th data point is $L_i$ accordingly. The data point with an extremum of the loss is likely to be mislabeled. Its bad impacts on model robustness should be handled with the M-estimator which is discussed later.

It is worth noting that, for technical implementation, the M-estimators share a similar idea with prior robust loss functions in tackling noisy labels, i.e., making the contributions of mislabeled data into optimization smaller (but not zero) for robustness enhancement. The difference between the M-estimators and robust loss functions is that the original M-estimators perform a subsequent reweighting process based on the magnitude of the loss, while robust loss functions output the loss in robust training directly.

\subsection{Representative M-estimators}\label{sec:2.5}
The robustness of statistical M-estimators has been carefully studied for several decades \cite{zhang2014novel}. One mainstream is to assign smaller weights to the data points with larger residuals \cite{catoni2012challenging} to make estimation results more robust. The reason for this is straightforward: the data points with large residuals are more likely to be \textit{outliers}. If we reduce their contributions to the optimization of the objective function, the results will be less influenced by outliers, and naturally will be more robust. We borrow some representative examples of robust M-estimators in this paper, which will be introduced as follows. The robust M-estimators used are denoted by $\Phi(\cdot)$. To make the description clearer, we will directly use the notations in learning with noisy labels, i.e., $\Phi(L)$. 

We compare assigned weights by robust M-estimators from an optimization viewpoint. That is to say, we compare the contributions to gradients brought by different examples, i.e., $\nabla\Phi(L)$. We exploit three robust M-estimators, i.e., Catoni's \cite{catoni2012challenging}, Log-sum Penalty \cite{candes2008enhancing}, and Welsch  \cite{liu2007correntropy}. For Welsch, we change $L^2$ to $L$ for weights assignments. The modified version is named Welsch+. The details of robust M-estimators used in this paper are provided in Table \ref{tab:m_estimator}.

\begin{table}[!t]
    \scriptsize
    \centering
    \begin{tabular}{lcc}
    \toprule
    M-estimators  & $\Phi(L)$ & $\nabla\Phi(L)$ \\
    \midrule
    CE \cite{mohri2018foundations} & $L$ & $\nabla L$\\
    \midrule
    Catoni's \cite{catoni2012challenging} & $\log\left(1+L+L^2/2\right)$& $\frac{1+L}{1+L+L^2/2}\nabla L$ \\
    \midrule
    Log-sum Penalty \cite{candes2008enhancing}&$\log(1+L/\epsilon)$ & $\frac{\epsilon}{\epsilon+L}\nabla L$\\
    \midrule
    Welsch+ \cite{liu2007correntropy} & $1-\exp\{-L/\alpha^2\}$ & $\frac{1}{\alpha^2}\exp\{-L/\alpha^2\}\nabla L$\\
    \bottomrule
    \end{tabular}
    \caption{\small{The definitions of used robust M-estimators.}}
    \vspace{-15pt}
    \label{tab:m_estimator}
\end{table}

\begin{figure}[!t]
\centering
\begin{subfigure}{.40\linewidth}
  \centering
  \includegraphics[width=1\linewidth]{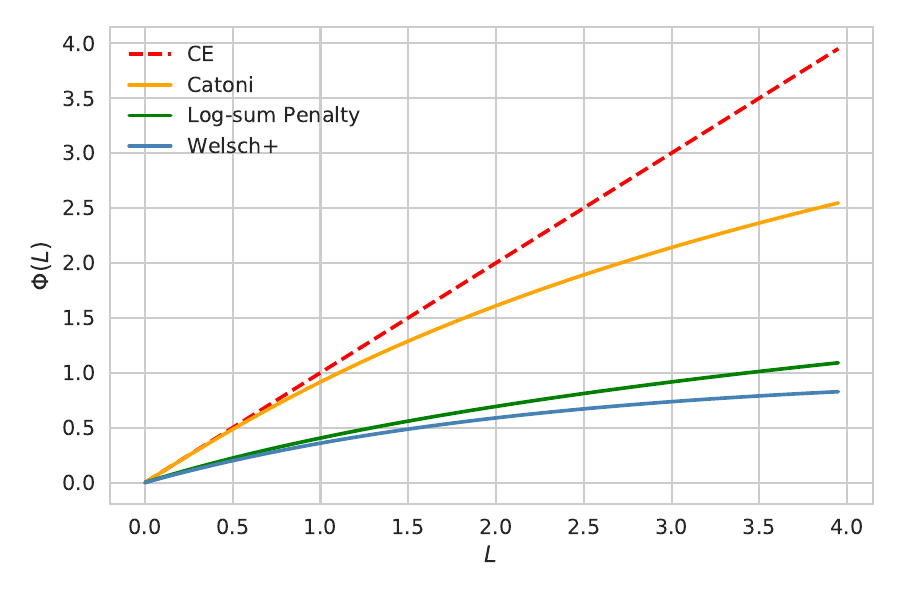} 
  \caption{}
\end{subfigure}
\hspace{5mm}
\begin{subfigure}{.40\linewidth}
  \centering
  \includegraphics[width=1\linewidth]{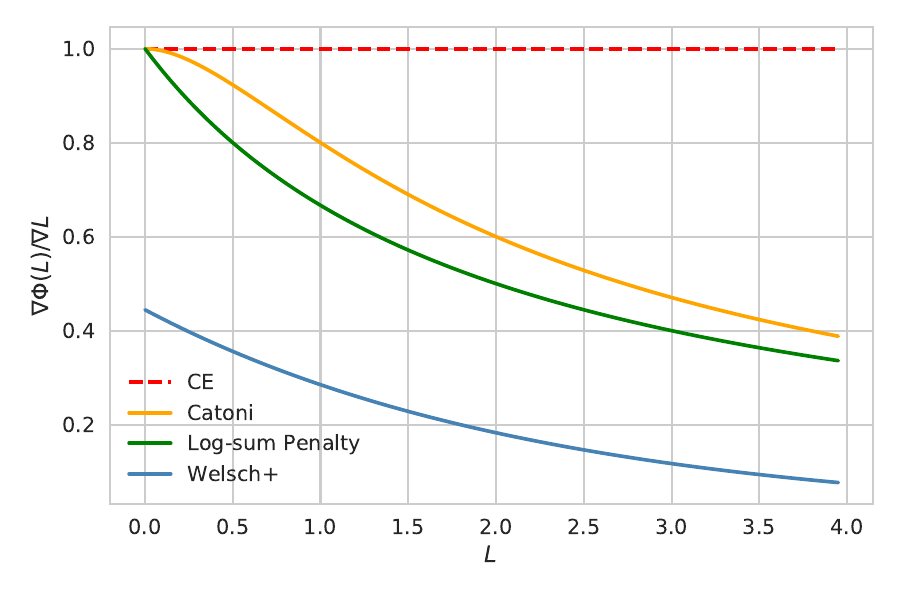} 
  \caption{}
\end{subfigure}
\vspace{-10pt}
\caption{\small{The illustrations of the used robust M-estimators, with $\epsilon=2$ and $\alpha=1.5$ for Log-sum Penalty and Welsch+. \textbf{(a):} $L$ \textit{vs} $\Phi(L)$. \textbf{(b):} $L$ \textit{vs} $\Phi(L)/\nabla L$.}}
\label{fig:m_estimators}
\end{figure}

Note that $\epsilon\in[1,+\infty)$ and $\alpha\in(0,+\infty)$  are parameters of Log-sum Penalty and Welsch+ respectively. For a better understanding of used robust M-estimators, we provide illustrations for $\Phi(L)$ and $\nabla\Phi(L)$, which are shown in Fig.~\ref{fig:m_estimators}. From the illustrations, we can see that robust M-estimators can change the behaviors of losses integrally. When the loss of an example is large, the example may be mislabeled. Robust M-estimators can reduce its loss value and its contribution to optimization during training.

\begin{figure}[!t]
\centering
\begin{subfigure}{.40\linewidth}
  \centering
  \includegraphics[width=1\linewidth]{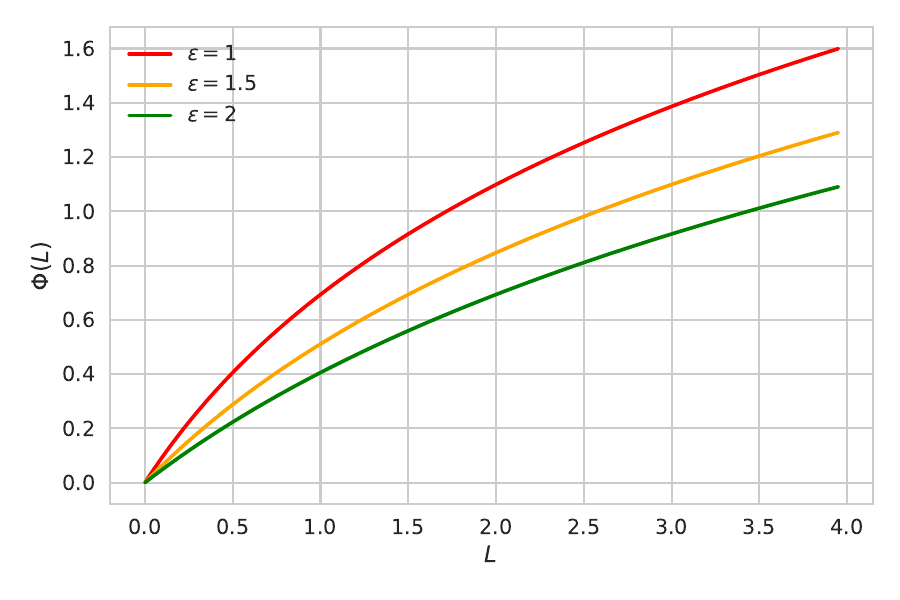} 
  \caption{}
\end{subfigure}
\hspace{5mm}
\begin{subfigure}{.40\linewidth}
  \centering
  \includegraphics[width=1\linewidth]{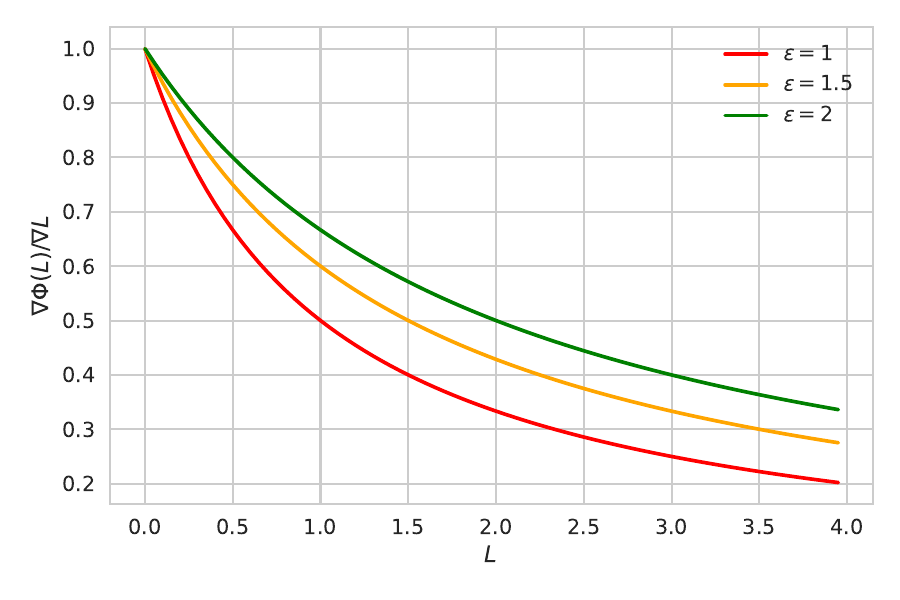} 
  \caption{}
\end{subfigure}
\vspace{5mm}
\begin{subfigure}{.40\linewidth}
  \centering
  \includegraphics[width=1\linewidth]{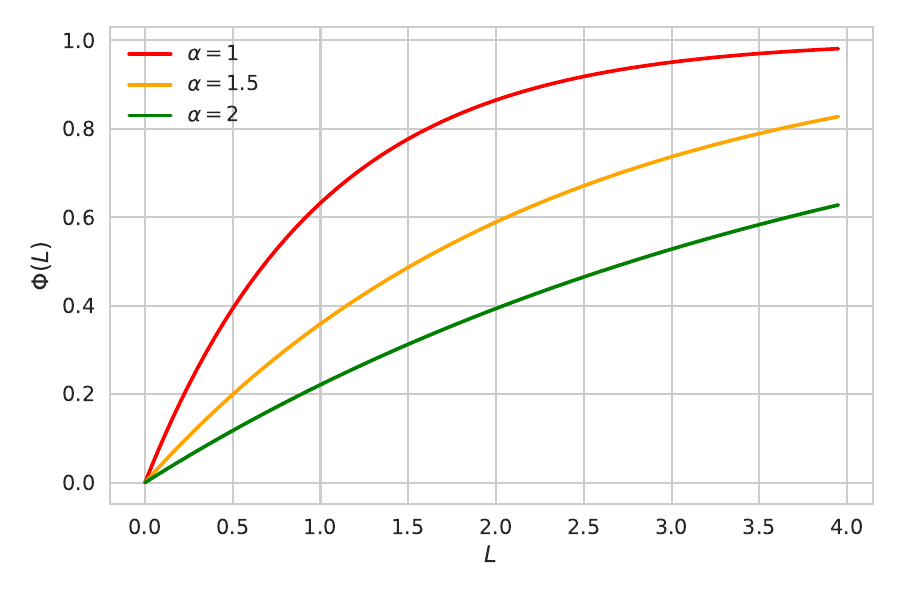} 
  \caption{}
\end{subfigure}
\hspace{5mm}
\begin{subfigure}{.40\linewidth}
  \centering
  \includegraphics[width=1\linewidth]{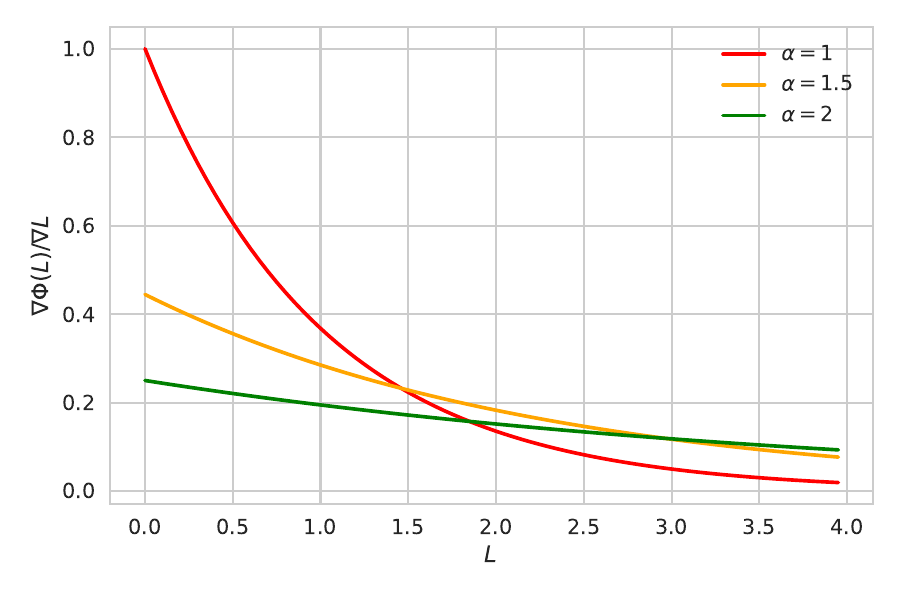} 
  \caption{}
\end{subfigure}
\vspace{-15pt}
\caption{\small{The illustrations of Log-sum Penalty and Welsch+ with different parameters, i.e., $\epsilon$ and $\alpha$. \textbf{(a):} Log-sum Penalty $L$ \textit{vs} $\Phi(L)$. \textbf{(b):} Log-sum Penalty $L$ \textit{vs} $\Phi(L)/\nabla L$. \textbf{(c):} Welsch+ $L$ \textit{vs} $\Phi(L)$. \textbf{(d):} Welsch+ $L$ \textit{vs} $\Phi(L)/\nabla L$.}}
\label{fig:m_estimators_parameters}
\end{figure}

Besides, the curves of Log-sum Penalty and Welsch+ with different parameters are plotted in Fig.~\ref{fig:m_estimators_parameters}. As can be seen, different parameters can control different penalties for large-loss examples. The choices of parameters of the estimators Log-sum Penalty and Welsch+, i.e., $\epsilon$ and $\alpha$, will be discussed in more detail later. 

\section{Methodology}\label{sec:3}
In this section, we formally present the proposed methods. We first propose how to perform truncation on the loss distribution automatically and achieve regularly truncated M-estimators (Section \ref{sec:3.1}). Afterward, the analyses of parameters of regularly truncated M-estimators are presented (Section \ref{sec:3.2}).

\subsection{The proposed algorithms}\label{sec:3.1}
We have discussed the mechanism of robust M-estimators. Nevertheless, we have two aspects that need to be considered carefully:
\begin{itemize}
    \item How to reduce the side effects of noisy labels in selected small-loss examples?
    \item How to make good use of large-losses examples to help generalization?
\end{itemize}
The first question can be answered immediately by using robust M-estimators on selected small-loss examples. For the second question, we need to think prudently. Specifically, large-losses examples may be clean as discussed. Moreover, even they may be mislabeled, their instances (e.g., images) still may be helpful \cite{xia2021instance}. However, due to the harmful influence of incorrect labels, large-loss examples should be used \textit{conservatively}. In this section, we formally present the proposed regularly truncated M-estimators to handle the mentioned problems at the same time.

\subsubsection{Truncated M-estimators} 
\begin{figure}[!t]
\centering
\begin{subfigure}{.4\linewidth}
  \centering
  \includegraphics[width=1\linewidth]{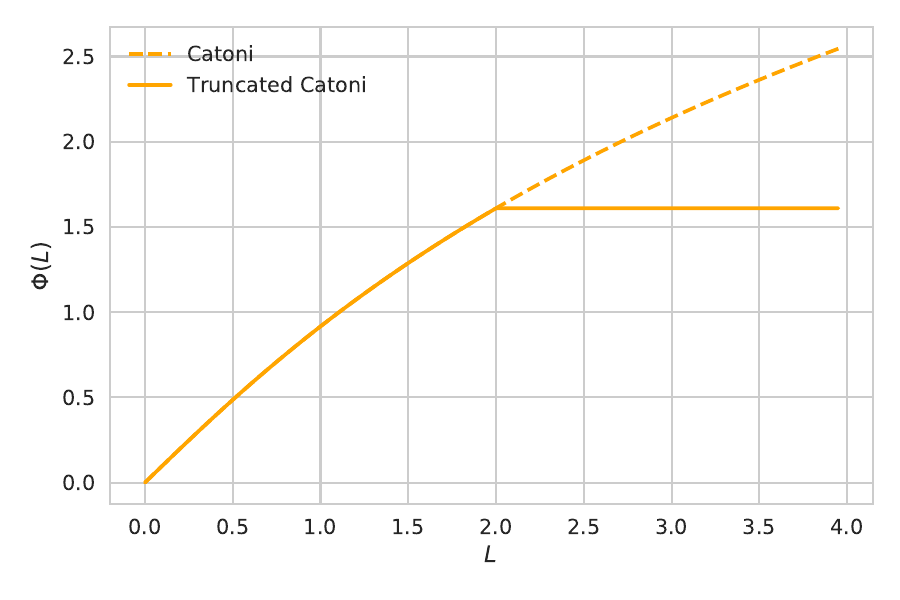} 
  \caption{}
  \label{fig:catoni_vs_truncated_catoni_a}
\end{subfigure}
\hspace{5mm}
\begin{subfigure}{.4\linewidth}
  \centering
  \includegraphics[width=1\linewidth]{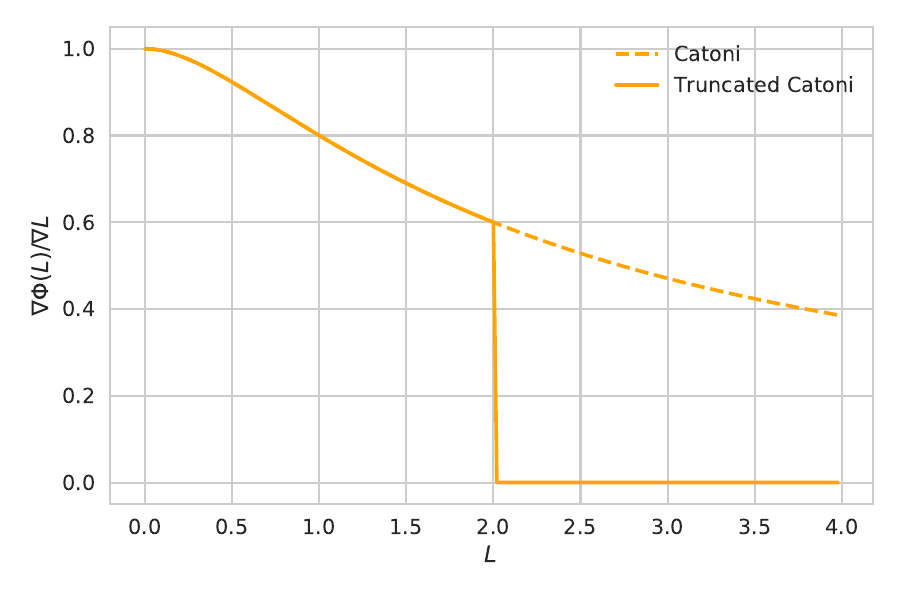} 
  \caption{}
  \label{fig:catoni_vs_truncated_catoni_b}
\end{subfigure}
\caption{\small{Truncated Catoni \textit{vs} Catoni. The truncation is performed at $\sigma=2$.}}
\label{fig:catoni_vs_truncated_catoni}
\end{figure}

To handle the first problem, i.e., using robust M-estimators on selected small-loss examples, we propose \textit{truncated M-estimators}. Namely, we perform truncation on the loss distribution. The truncation divides all examples into small-loss ones and large-loss ones. We then can employ robust M-estimators to reduce the side effects of noisy labels in selected small-loss examples. By using the M-estimators Catoni's, the truncated M-estimators are defined as follows: 
\begin{equation}
    \Phi^{T}(L)=
    \begin{cases}
    \log(1+L+L^2/2)& L\leq\sigma\\
    \log(1+\sigma+\sigma^2/2)& L\textgreater\sigma
\end{cases}
\end{equation}
where $\sigma\textgreater 0$ is a hyperparameter related to the loss distribution to control the truncated point (or threshold). Other truncated robust M-estimators are provided in Table \ref{tab:t_m_estimator}. The comparison between truncated Catoni's and Catoni's is provided in Fig.~\ref{fig:catoni_vs_truncated_catoni}. As can be seen, truncated Catoni's reserves the nice properties so that it can assign different weights to selected small-loss examples to relieve the influence of noisy labels (Fig.~\ref{fig:catoni_vs_truncated_catoni_a}). Meanwhile, in fact, it directly removes large-loss examples from training since such examples have no contribution to optimization (Fig.~\ref{fig:catoni_vs_truncated_catoni_b}). 

\begin{table*}[!h]
    \centering
    \scriptsize
    \begin{tabular}{lcc}
    \toprule
    Truncated M-estimators  & $\Phi^T(L)$ & $\nabla\Phi^T(L)$ \\
    \midrule
    Truncated Catoni's  &$\begin{cases}
    \log(1+L+L^2/2)& L\leq\sigma\\
    \log(1+\sigma+\sigma^2/2)& L\textgreater\sigma
    \end{cases}$& $\begin{cases}
    \frac{1+L}{1+L+L^2/2}\nabla L& L\leq\sigma\\
    0& L\textgreater\sigma
    \end{cases}$ \\
    \midrule
    Truncated Log-sum Penalty & $\begin{cases}
    \log(1+L/\epsilon) & L\leq\sigma\\
    \log(1+\sigma/\epsilon) & L\textgreater\sigma
    \end{cases}$ & $\begin{cases}
    \frac{\epsilon}{\epsilon+L}\nabla L & L\leq\sigma\\
    0 & L\textgreater\sigma
    \end{cases}$\\
    \midrule
    Truncated Welsch+  & $\begin{cases}
    1-\exp\{-L/\alpha^2\} & L\leq\sigma\\
    1-\exp\{-\sigma/\alpha^2\} & L\textgreater\sigma
    \end{cases}$ & $\begin{cases}
    \frac{1}{\alpha^2}\exp\{-L/\alpha^2\}\nabla L & L\leq\sigma\\
    0 & L\textgreater\sigma
    \end{cases}$ \\
    \bottomrule
    \end{tabular}
    \caption{\small{The definitions of proposed truncated M-estimators.}}
    \label{tab:t_m_estimator}
\end{table*}

\noindent\textbf{Theoretical properties.} We discuss the theoretical properties of the proposed truncated M-estimators. We demonstrate that they are noise-tolerant. That is, the minimizers of the risk under the truncated M-estimators with noisy
labels would be the same as those with noise-free labels.

\begin{lemma}\label{lemma1}
    In a multi-class classification problem, the truncated M-estimators are noise-tolerant under symmetric (or uniform) label noise, if $c_2-c_1-k\Delta(\psi,f)>0$ and the noise rate $\eta<\frac{(1-k)\Delta(\psi,f)}{c_2-c_1-k\Delta(\psi,f)}$. Here $c_1$ and $c_2$ denote the lower and upper bounds of the sum of the losses obtained by predictions on all classes, and $\Delta(\psi,f)$ is the gap between the clean risk w.r.t. $(\psi,f)$ and the minimum clean risk brought by the global minimizer about $f$. 
\end{lemma}
Note that inspired by \cite{ghosh2017robust}, the theoretical analysis can be extended to simple non-uniform noise under some conditions. Due to the limited page of the main paper, more background knowledge and detailed proofs of Lemma~\ref{lemma1} are provided in Appendix~\textcolor{red}{A}. 

\begin{remark}
    The philosophy of noise tolerance of the truncated M-estimators is similar to the noise tolerance of some other robust loss functions that make the value of the loss sum \textit{bounded}. For example, \cite{ghosh2017robust} considers the value of loss sum to be a constant $C$. Besides, \cite{ma2020normalized} considers it to be 1. Differently, our truncated M-estimators employ a truncation mechanism to restrain the value of the loss sum, since the loss that is larger than $\sigma$ is limited to be fixed, e.g., $\log(1+\sigma+\sigma^2/2)$ for truncated Catoni's. Based on this, the paper provides a new perspective to make one loss function robust to noisy labels. 
\end{remark}

\subsubsection{Regularly truncated M-estimators.} As truncated M-estimators cannot handle the second problem, to handle the first and second problem at the same time, we further propose \textit{regularly} truncated M-estimators. Here, the term ``regularly'' means that we alternately exploit truncated robust M-estimators and original robust M-estimators. Formally, we define 
\begin{equation}\label{eq:regularly}
    \Phi^R(L)=\mathbbm{1}_{\{T \ \text{mod} \ R \ \neq \ 0\}}\Phi^T(L) + \mathbbm{1}_{\{T \ \text{mod} \ R \ = \ 0\}}\Phi(L),
\end{equation}
where $R\in\mathbbm{N}_{+}$ is the hyperparameter about the frequency of using different kinds of robust M-estimators. Apparently, if the value of $R$ is large, the large-loss examples will be involved in optimization \textit{infrequently}, i.e., truncated M-estimators are often employed to perform reweighting on selected small-loss examples. Oppositely, if the value of $R$ is small, large-loss examples will be involved in optimization \textit{more frequently}. Distinctly, we need to choose a suitable $R$ to achieve a great balance between truncated ones and original ones, which can be determined with a clean or noisy validation set. We will discuss this in Section \ref{sec:4}.
\subsection{Parameters analyses}\label{sec:3.2}
For the truncated M-estimators, we have two types of parameters that need to be determined. The first type of parameter is the truncation point $\sigma$. The second type of parameter is the intrinsic parameter of M-estimators, i.e., $\epsilon$ and $\alpha$. In this subsection, we discuss how to determine them.

We discuss how to determine $\sigma$. We borrow the ``three-sigma'' rule from the probability theory
and statistics \cite{durrett2019probability} rather than estimating the noise rate, since the noise rate is hard to be estimated in some cases \cite{xia2020part}. The ``three-sigma'' rule has been verified to be effective to remove underlying outliers \cite{guan2017truncated}.  Specifically, let $\mathcal{L}\subset\mathbbm{R}^n$ denote the losses of all training examples for each epoch. We first find the subset $\Gamma=\{0\leq L_i\leq M|L_i\in\mathcal{L}\}$, where $M$ represents the \textit{median} of the set $\mathcal{L}$. Then we calculate the mean $\mu$ and standard deviation $\delta$ of the losses in $\Gamma$. Finally, we set the threshold $\sigma=\mu+3\delta$. The threshold $\sigma$ can be updated at every epoch according to the loss distribution. In Section \ref{sec:4}, we will provide the experimental results for the justification of determining $\sigma$.  

We then discuss how to determine $\epsilon$ and $\alpha$ during training. The parameter determination problem has been studied for several decades \cite{nagy2006parameter}. There are two main ways to determine this. On the one hand, we can empirically set the parameter in a reasonable range. On the other hand, we can assume the distribution of data to help determine. We follow both ways for the determination of  $\epsilon$ and $\alpha$. For the first way, we simply set $\epsilon=\alpha=1$. For the second way, we assume that the outputs of M-estimators for selected small loss examples (denoted by $\Gamma'$) obey a \textit{Gaussian} distribution. More specifically, we calculate the mean $\mu'$ and standard deviation $\delta'$. Then we tune $\epsilon$ (resp. $\alpha$) to make that the distribution of $\Gamma'$ is closer to $\mathcal{N}(\mu', \delta'^2)$. 

The overall procedure of the proposed method is provided in Algorithm \ref{alg:ours}. As can be seen, in each epoch, we first determine the needed parameters (Step 4). Then when the mini-batch data is formed (Step 6), we use the proposed regularly truncated M-estimators on it (Step 7). The proposed method is easy to follow and can keep an \textit{end-to-end} manner.  

\begin{algorithm}[!t]
\caption{The procedure of regularly truncated M-estimators for learning with noisy labels.}
\begin{algorithmic}[1]
	    \STATE \textbf{Input}: initialized classifier $f$, epoch $T_{\max}$, iteration $t_{\max}$, and the frequency $R$.
		\FOR{$T = 0, \dots, T_{\max}-1$}
		\STATE \textbf{Shuffle} the training dataset $S$;
		\STATE \textbf{Determine} the truncation parameter $\sigma$ and intrinsic parameters $\epsilon$/$\alpha$ as discussed in Section \ref{sec:3.2};
		\FOR{$t = 0, \dots, t_{\max}-1$}
		\STATE \textbf{Draw} a mini-batch $\bar{S}$ from $S$;
		\STATE \textbf{Perform} regularly truncated M-estimators on $\bar{S}$ with Eq.~(\ref{eq:regularly});
		\STATE \textbf{Update} classifier parameters;
		\ENDFOR
		\ENDFOR
		\STATE \textbf{Output}: trained classifier $f$.
	\end{algorithmic}
	\label{alg:ours}
\end{algorithm}

\section{Experiments}\label{sec:4}
In this section, we experimentally explore both the robustness and effectiveness of the proposed method. We first introduce the methods for comparison in the experiments (Section \ref{sec:4.1}). We then introduce the details of the experiments on synthetic datasets (Section \ref{sec:4.2}). The experiments on real-world datasets are finally presented (Section \ref{sec:4.3}). 

\subsection{Comparison methods}\label{sec:4.1}
We compare our method with multiple baselines, which include broad types of advanced approaches for combating noisy labels. The overview and publication locations of the baselines are summarized as follows.
\begin{itemize}
    \item \textbf{Robust loss functions}. (1). \underline{APL}~(ICML 2020)~\cite{ma2020normalized}, which combines two mutually reinforcing robust loss functions. (2). \underline{PCE}~(ICLR 2020)~\cite{menon2020can}, which boosts the standard cross entropy loss with a partial trick. The tuning parameter of PCE is set to 2 in experiments. (3) AUL~(T-PAMI 2023)~\cite{zhou2023asymmetric}, which are tailored to satisfy the Bayes-optimal condition and thus are robust to noisy labels under some conditions. (4) CELC~(ICML 2023)~\cite{wei2023mitigating}, which induces a loss bound at the logit level, thus enhancing the noise robustness of the softmax cross entropy loss. 
    \item \textbf{Loss correction}. (1). \underline{Revision}~(NeurIPS 2019)~\cite{xia2019anchor}, which introduces a slack variable to revise the noise transition matrix, leading to a better classifier. (2). Identifiability~(ICML 2023)~\cite{liu2023identifiability}, which improves the estimation of the transition matrix using properly disentangled features.
    \item \textbf{Label correction}. (1). \underline{Joint}~(CVPR 2018)~\cite{tanaka2018joint}, which jointly optimizes the network parameters and the sample labels. The hyperparameters $\alpha$ and $\beta$ for Joint are set to 1.2 and 0.8 respectively.
    \item \textbf{Sample selection}. (1). \underline{Co-teaching}~(NeurIPS 2018)~\cite{han2018co}, which trains two networks simultaneously and cross-updates parameters of peer networks. (2). \underline{SIGUA}~(ICML 2020)~\cite{han2020sigua}, which exploits stochastic integrated gradient underweighted ascent to handle noisy labels. We use self-teach SIGUA in this paper. (3). \underline{Co-Dis}~(ICCV 2023)~\cite{xia2023co}, which selects possibly clean data that simultaneously have high-variance prediction probabilities between two networks. For these methods, we reserve their hyperparameter and optimization settings for selecting clean examples during training. Besides, we use an estimated noise rate \cite{liu2016classification} for them to ensure a fair comparison. 
\end{itemize}

As a simple baseline, we compare our method with the standard deep network that directly trains on noisy
datasets by using the softmax cross entropy loss function (abbreviated as \underline{CE}). Note that we do not directly compare the proposed method with some state-of-the-art methods, e.g., SELF \cite{nguyen2020self} and DivideMix \cite{li2020dividemix}. It is because their proposed methods are aggregations of multiple techniques, e.g., Mixup \cite{zhang2017mixup,li2022selective}, soft labels \cite{reed2014training}, and semi-supervised learning \cite{berthelot2019mixmatch}. We mainly focus on sample selection in learning with noisy labels. Therefore, the comparison is not fair. To make a fair comparison, we combine our method with semi-supervised learning and self-supervised learning to combat noisy labels. More details will be shown in Sections~\ref{sec:combine_with_ssl} and \ref{sec:combine_with_self}.

\subsection{Experiments on simulated noisy datasets}\label{sec:4.2}
\subsubsection{Experimental setup}
\textbf{Datasets.} We verify the effectiveness of our methods on the manually corrupted version of the following datasets: \textit{MNIST} \cite{LeCunmnist}, \textit{SVHN} \cite{netzer2011svhn}, \textit{CIFAR-10} \cite{krizhevsky2009learning}, \textit{CIFAR-100} \cite{krizhevsky2009learning}, and \textit{NEWS} \cite{lang1995newsweeder}, because these datasets are popularly used for the evaluation of learning with noisy labels in the literature \cite{han2018co,yu2019does,wu2020class2simi,lee2019robust}. For \textit{NEWS}, we borrowed
the pre-trained word embeddings from GloVe \cite{pennington2014glove}. The important statistics of the used synthetic datasets are summarized in Table \ref{tab:syn_dataset}. 
\begin{table}[!t]
    \centering
    \tiny
    \begin{tabular}{cccccc}
    \toprule	 
         Datasets & Type &\# Of training&\# Of testing&\# Of class & Size\\
         \midrule
         \textit{MNIST} & image & 60,000 & 10,000 & 10 & 28$\times$28$\times$1\\
        \midrule
         \textit{SVHN}  & image & 73,257 & 26,032 & 10 & 32$\times$32$\times$3\\
         \midrule
         \textit{CIFAR-10} & image & 50,000 & 10,000 & 10 & 32$\times$32$\times$3\\
         \midrule
         \textit{CIFAR-100} & image & 50,000 & 10,000 & 100 & 32$\times$32$\times$3\\
         \midrule
         \textit{NEWS} & text & 11,314 & 7,532 & 20 & 300-D\\
    \bottomrule	 	 
    \end{tabular}
    \caption{\small{Summary of simulated noisy datasets used in the experiments.}}
    \label{tab:syn_dataset}
    \vspace{-15pt}
\end{table}

\begin{table*}[!t]
    \centering
    \scriptsize
    \begin{tabular}{ll|cccccc}
    \toprule
    Datasets & Methods / Noise &  Sym.-30\% & Sym.-50\% & Pair.-30\% & Pair.-45\% & Ins.-30\% & Ins.-50\% \\\midrule
    \multirow{14}{*}{\textit{MNIST}} & CE  & 96.29 $\pm$ 0.04 & 94.85 $\pm$ 0.12 & 95.15 $\pm$ 0.06 & 93.92 $\pm$ 0.39 & 95.87 $\pm$ 0.12 & 81.52 $\pm$ 5.52 \\
    & APL & 96.22 $\pm$ 0.08 & 95.86 $\pm$ 0.26 & 96.28 $\pm$ 0.09 & 92.40 $\pm$ 0.69 & 90.07 $\pm$ 3.91 & 72.22 $\pm$ 15.36\\
    & PCE & 95.77 $\pm$ 0.62 & 95.07 $\pm$ 0.18 & 96.04 $\pm$ 0.17 & 93.92 $\pm$ 1.04 & 96.02 $\pm$ 0.47 & 78.93 $\pm$ 4.07\\
    & AUL & 94.07 $\pm$ 0.15 & 79.80 $\pm$ 3.64 & 60.42 $\pm$ 3.91 & 60.17 $\pm$ 3.69 & 92.16 $\pm$ 0.76 & 73.55 $\pm$ 7.17\\
    & CELC & 96.19 $\pm$ 0.11 & 95.35 $\pm$ 0.35 & 96.19 $\pm$ 0.98 & \textbf{95.84 $\pm$ 1.23} & 96.15 $\pm$ 1.38 & 89.15 $\pm$ 3.88 \\
    & Revision & 96.47 $\pm$ 0.17 & 95.79 $\pm$ 0.24 & 96.08 $\pm$ 0.14 & 94.19 $\pm$ 0.93 & \underline{\textbf{96.49 $\pm$ 0.24}} & 85.47 $\pm$ 3.04\\
    & Identifiability & \underline{\textbf{97.09 $\pm$ 0.35}} & 95.86 $\pm$ 0.69 & \underline{\textbf{97.52 $\pm$ 0.47}} & \underline{\textbf{97.48 $\pm$ 0.81}} & \textbf{96.28 $\pm$ 0.56} & 88.44 $\pm$ 1.04 \\
    & Joint & 96.26 $\pm$ 0.15 & 94.09 $\pm$ 0.47 & 94.02 $\pm$ 0.19 & 93.78 $\pm$ 0.92 & 96.03 $\pm$ 0.15 & 86.49 $\pm$ 4.15\\
    & Co-teaching & 96.04 $\pm$ 0.07 & 95.07 $\pm$ 0.24 & 96.09 $\pm$ 0.14 & 94.37 $\pm$ 0.58 & 94.53 $\pm$ 0.29 & 87.52 $\pm$ 2.44\\
    & SIGUA & 95.37 $\pm$ 0.93 & 95.07 $\pm$ 0.84 & 94.73 $\pm$ 0.39 & 90.04 $\pm$ 1.83 & 93.14 $\pm$ 1.29 & 80.47 $\pm$ 9.39\\
    & Co-Dis & 96.48~$\pm$~0.15 & 95.37~$\pm$~0.27 & 96.21~$\pm$~0.14 & 94.20~$\pm$~1.05 & 95.55~$\pm$~1.03 & 90.33~$\pm$~1.11\\\cmidrule{2-8}
    & RT-Catoni's & \textbf{96.56 $\pm$ 0.04} & \underline{\textbf{96.33 $\pm$  0.17}} & 96.87 $\pm$ 0.12 & 95.72 $\pm$  0.82 & 96.08 $\pm$  0.18 & \textbf{93.25 $\pm$  0.66}\\
    & RT-Log-sum & \textbf{96.53 $\pm$ 0.12} & \textbf{96.21 $\pm$ 0.19} & \textbf{96.92 $\pm$ 0.09} & 95.70 $\pm$ 0.82 & 96.06 $\pm$ 0.14 & \underline{\textbf{93.77 $\pm$ 0.72}}\\
    & RT-Welsch+ & 96.44 $\pm$ 0.06 & \textbf{95.90 $\pm$ 0.25} & \textbf{96.96 $\pm$ 0.04} & \textbf{96.56 $\pm$ 0.14} & \textbf{96.26 $\pm$ 0.05} & \textbf{93.02 $\pm$ 3.53}\\
    \bottomrule
    \bottomrule
    \multirow{14}{*}{\textit{SVHN}} & CE  & 92.75 $\pm$ 0.31 & 90.63 $\pm$ 0.71 & 93.82~$\pm$~0.13 & 70.95~$\pm$~2.38 & 93.31~$\pm$~0.37 & 63.16~$\pm$~8.12 \\
    & APL & 93.82~$\pm$~0.19 & 91.34~$\pm$~0.37 & 94.69~$\pm$~0.19 & 86.77~$\pm$~0.41 & 94.01~$\pm$~0.36 & \textbf{67.61~$\pm$~9.80}\\
    & PCE & 93.81~$\pm$~0.64 & 90.73~$\pm$~0.19 & 94.24~$\pm$~0.61 & 87.16~$\pm$~1.14 & 93.31~$\pm$~0.52 & 63.10~$\pm$~7.26\\
    & AUL & 94.44~$\pm$~0.52 & 92.75~$\pm$~0.39 & 94.80~$\pm$~1.20 & 82.77~$\pm$~2.61 & 94.35~$\pm$~0.16 & 64.33~$\pm$~5.23\\
    & CELC & 95.06 $\pm$ 0.41 & 92.51 $\pm$ 0.89 & 94.36 $\pm$ 0.52 & 88.78 $\pm$ 1.37 & 94.16 $\pm$ 0.61 & 66.12 $\pm$ 3.44 \\
    & Revision & 94.20~$\pm$~0.22 & 94.06~$\pm$~0.19 & 94.78~$\pm$~0.30 & 81.36~$\pm$~1.82 & 94.53~$\pm$~0.57 & \textbf{67.21~$\pm$~4.94}\\
    & Identifiability & 93.18 $\pm$ 0.71 & 92.06 $\pm$ 1.33 & 92.66 $\pm$ 0.95 & 85.56 $\pm$ 1.40 &  92.01 $\pm$ 1.90 & 66.04 $\pm$ 5.71\\
    & Joint & 93.37~$\pm$~0.27 & 92.11~$\pm$~0.63 & 93.79~$\pm$~0.29 & 75.86~$\pm$~1.73 & 94.63~$\pm$~0.82 & 62.19~$\pm$~6.95\\
    & Co-teaching & 93.79~$\pm$~0.67 & 92.63~$\pm$~0.43 & 94.15~$\pm$~0.62 & 88.36~$\pm$~0.95 & 93.14~$\pm$~0.12 & 61.55~$\pm$~8.75\\
    & SIGUA & 94.04~$\pm$~1.31 & 90.55~$\pm$~2.44 & 92.19~$\pm$~1.21 & 74.44~$\pm$~5.72 & 92.66~$\pm$~0.61 & 57.92~$\pm$~11.68\\
    & Co-Dis & 94.77~$\pm$~0.58 & 93.02~$\pm$~0.82 & 94.78~$\pm$~0.29 & 90.06~$\pm$~1.03 & 93.77~$\pm$~0.29 & 63.32~$\pm$~8.80\\\cmidrule{2-8}
    & RT-Catoni's & \underline{\textbf{95.54~$\pm$~0.17}} & \underline{\textbf{94.70~$\pm$~0.20}} & \underline{\textbf{95.29~$\pm$~0.10}} & \textbf{92.69~$\pm$~0.83} & \textbf{94.69~$\pm$~0.24} & \underline{\textbf{68.00~$\pm$~13.15}}\\
    & RT-Log-sum & \textbf{95.51~$\pm$~0.15} & \textbf{94.54~$\pm$~0.21} & \textbf{95.25~$\pm$~0.22} & \textbf{91.59~$\pm$~2.17} & \textbf{94.92~$\pm$~0.19} & 66.96~$\pm$~12.57\\
    & RT-Welsch+ & \textbf{95.44~$\pm$~0.08} & \textbf{94.47~$\pm$~0.16} & \textbf{95.16~$\pm$~0.36} & \underline{\textbf{92.89~$\pm$~0.71}} & \underline{\textbf{94.99~$\pm$~0.23}} & 61.60~$\pm$~15.63\\
    \bottomrule
    \bottomrule
    \multirow{14}{*}{\textit{CIFAR-10}} & CE  & 82.67~$\pm$~0.48 & 76.01~$\pm$~1.43 & 84.97~$\pm$~1.04 & 61.76~$\pm$~4.53 & 83.15~$\pm$~0.55 & 54.29~$\pm$~3.90\\
    & APL & 85.54~$\pm$~0.51 & 78.36~$\pm$~0.47 & 85.40~$\pm$~0.14 & 80.84~$\pm$~0.72 & 77.57~$\pm$~0.15 & 39.45~$\pm$~6.51\\
    & PCE & 86.12~$\pm$~0.85 & 74.03~$\pm$~4.96 & 85.03~$\pm$~0.77 & 65.08~$\pm$~3.41 & 85.64~$\pm$~0.72 & 64.82~$\pm$~4.13\\
    & AUL & 88.09~$\pm$~0.78 & 82.81~$\pm$~1.16 & 71.34~$\pm$~1.91 & 56.80~$\pm$~2.69 & 86.35~$\pm$~0.90 & 60.75~$\pm$~3.77\\
    & CELC & \textbf{89.46 $\pm$ 2.13} & 85.08 $\pm$ 3.95 & 89.77 $\pm$ 2.56 & \textbf{85.72 $\pm$ 4.52} & 86.67 $\pm$ 1.47 & 61.85 $\pm$ 4.98 \\
    & Revision & 88.39~$\pm$~0.38 & 83.40~$\pm$~0.65 & \textbf{90.70~$\pm$~0.47} & 83.61~$\pm$~1.06 & 89.07~$\pm$~0.35 & \textbf{66.93~$\pm$~4.14}\\
    & Identifiability  & 87.12 $\pm$ 1.69 & 83.43 $\pm$ 2.11 & 86.45 $\pm$ 1.93 & 83.65 $\pm$ 2.46 & 80.47 $\pm$ 1.54 & 55.25 $\pm$ 3.78\\
    & Joint & 89.34~$\pm$~0.52 & 85.06~$\pm$~0.29 & 89.75~$\pm$~0.63 & 80.52~$\pm$~1.90 & 88.41~$\pm$~1.02 & 64.12~$\pm$~3.89\\
    & Co-teaching & 88.93~$\pm$~0.56 & 85.73~$\pm$~0.12 & 88.72~$\pm$~0.61 & 84.19~$\pm$~0.68 & 87.07~$\pm$~0.35 & 60.09~$\pm$~3.31\\
    & SIGUA & 83.19~$\pm$~1.26 & 77.92~$\pm$~3.11 & 83.93~$\pm$~0.49 & 70.39~$\pm$~1.94 & 82.90~$\pm$~2.00 & 30.95~$\pm$~9.70\\
    & Co-Dis & 89.20~$\pm$~0.13 & 85.36~$\pm$~0.94 & 89.20~$\pm$~0.37 & \textbf{85.02~$\pm$~1.33} & 87.13~$\pm$~0.25 & 62.77~$\pm$~3.90\\\cmidrule{2-8}
    & RT-Catoni's & 89.39~$\pm$~0.28 & \underline{\textbf{87.00~$\pm$~0.08}} & \underline{\textbf{90.83~$\pm$~0.20}} & \underline{\textbf{86.57~$\pm$0.92}} & \textbf{89.34~$\pm$~0.32} & \underline{\textbf{69.77~$\pm$~2.14}} \\
    & RT-Log-sum & \textbf{89.60~$\pm$~0.44} & \textbf{87.41~$\pm$~0.30} & \textbf{90.49~$\pm$~0.12} &83.60~$\pm$~1.38 & \underline{\textbf{89.65~$\pm$~0.88}} & \textbf{68.97~$\pm$~3.82}\\
    & RT-Welsch+ & \underline{\textbf{90.65~$\pm$~0.22}} & \textbf{86.60~$\pm$~0.51} & 90.15~$\pm$~0.38 & 77.29~$\pm$~6.52 & \textbf{89.56~$\pm$~0.62} & 60.86~$\pm$~10.60\\
    \bottomrule
    \bottomrule
    \multirow{14}{*}{\textit{CIFAR-100}} & CE  & 51.25~$\pm$~0.50 & 40.28~$\pm$~0.53 & 51.71~$\pm$~0.63 & 38.54~$\pm$~0.53 & 52.02~$\pm$~0.44 & 36.35~$\pm$~0.87 \\
    & APL & 55.78~$\pm$~0.91 & 46.96~$\pm$~0.81 & 56.34~$\pm$~0.68 & 49.55~$\pm$~1.05 & 43.30~$\pm$~1.57 & 29.01~$\pm$~0.09\\
    & PCE & 58.84~$\pm$~1.32 & 42.63~$\pm$~2.02 & 54.23~$\pm$~1.76 & 41.05~$\pm$~2.83 & 55.72~$\pm$~1.96 & 38.72~$\pm$~3.01\\
    & AUL & 69.89~$\pm$~0.21 & 60.00~$\pm$~0.40 & 64.96~$\pm$~0.55 & 39.37~$\pm$~1.61 & 67.75~$\pm$~1.84 & 40.27~$\pm$~1.76 \\
    & CELC & 67.96 $\pm$ 1.88 & 60.71 $\pm$ 2.39 & 67.96 $\pm$ 2.10 & \textbf{52.53 $\pm$ 3.17} & 66.25 $\pm$ 1.93 & 47.52 $\pm$ 3.93 \\
    & Revision & 62.97~$\pm$~0.46 & 43.60~$\pm$~0.94 & 60.09~$\pm$~1.21 & 49.33~$\pm$~1.10 & 56.46~$\pm$~1.45 & 40.78~$\pm$~1.75\\
    & Identifiability & 50.53 $\pm$ 1.52 & 34.87 $\pm$ 2.36 & 52.88 $\pm$ 1.15 & 38.16 $\pm$ 2.68 & 52.48 $\pm$ 1.93 & 36.72 $\pm$ 3.10 \\
    & Joint & 63.69~$\pm$~0.84 & 55.62~$\pm$~1.68 & 65.11~$\pm$~1.79 & 49.77~$\pm$~1.15 & 64.15~$\pm$~1.11 & 45.47~$\pm$~2.73\\
    & Co-teaching & 59.49~$\pm$~0.36 & 52.19~$\pm$~1.42 & 54.92~$\pm$~2.84 & 47.53~$\pm$~1.39 & 56.71~$\pm$~1.26 & 42.09~$\pm$~1.73\\
    & SIGUA & 54.22~$\pm$~0.90 & 50.64~$\pm$~3.92 & 47.92~$\pm$~2.93 & 39.92~$\pm$~2.33 & 53.19~$\pm$~2.64 & 38.50~$\pm$~1.69\\
    & Co-Dis & 64.02~$\pm$~1.37 & 54.55~$\pm$~2.06 & 58.72~$\pm$~2.11 & \textbf{50.02~$\pm$~2.80} & 59.15~$\pm$~1.92 & 43.38~$\pm$~1.25\\\cmidrule{2-8}
    & RT-Catoni's & \textbf{70.04~$\pm$~0.28} & \textbf{64.87~$\pm$~0.52} & \underline{\textbf{71.75~$\pm$~0.33}} & \textbf{50.02~$\pm$~0.95} & \underline{\textbf{71.66~$\pm$~0.53}} & \textbf{53.97~$\pm$~0.45}\\
    & RT-Log-sum & \underline{\textbf{70.30~$\pm$~0.32}} & \underline{\textbf{65.20~$\pm$~0.44}} & \textbf{71.68~$\pm$~0.18} & 48.16~$\pm$~1.26 & \textbf{71.22~$\pm$~0.50} & \textbf{54.09~$\pm$~0.37}\\
    & RT-Welsch+ & \textbf{69.17~$\pm$~0.60} & \textbf{57.63~$\pm$~0.92} & \textbf{69.34~$\pm$~0.50} & \underline{\textbf{54.00~$\pm$~1.50}} & \textbf{69.22~$\pm$~0.17} & \underline{\textbf{56.44~$\pm$~1.78}}\\
    \bottomrule
    \bottomrule
    \multirow{14}{*}{\textit{NEWS}} & CE  & 43.16~$\pm$~1.95 & 32.92~$\pm$~0.86 & 42.86~$\pm$~1.06 & 28.33~$\pm$~3.58 & 44.08~$\pm$~1.70 & 30.06~$\pm$~7.92 \\
    & APL & 54.04~$\pm$~1.09 & 45.12~$\pm$~2.17 & 51.98~$\pm$~0.27 & 36.86~$\pm$~2.31 & 52.18~$\pm$~0.63 & 44.82~$\pm$~3.61\\
    & PCE & 55.12~$\pm$~0.94 & 49.77~$\pm$~0.32 & 54.17~$\pm$~0.98 & 37.92~$\pm$~2.02 & 54.37~$\pm$~0.95 & 46.14~$\pm$~1.29\\
    & AUL & 53.77~$\pm$~0.25 & 48.78~$\pm$~1.62 & 53.72~$\pm$~1.77 & 39.23~$\pm$~1.06 & 55.19~$\pm$~1.09 & 47.73~$\pm$~2.11\\
    & CELC & 52.15~$\pm$~0.86 & 47.25~$\pm$~1.00 & 52.50~$\pm$~0.84 & 38.10~$\pm$~1.06 & 53.70~$\pm$~1.81 & 47.00~$\pm$~2.06\\
    & Revision & 55.19~$\pm$~0.63 & 50.65~$\pm$~0.97 & 53.77~$\pm$~0.64 & 38.91~$\pm$~1.38 & 53.29~$\pm$~0.62 & 46.37~$\pm$~2.94\\
    & Identifiability & 53.65 $\pm$ 1.65 & 50.84 $\pm$ 2.27 &  53.16  $\pm$ 1.95 & 39.16 $\pm$ 2.62 & 52.35 $\pm$ 1.92 & 44.87 $\pm$ 3.92 \\
    & Joint & 53.15~$\pm$~0.92 & 48.77~$\pm$~1.47 & 51.90~$\pm$~1.35 & 33.29~$\pm$~3.45 & 52.92~$\pm$~0.64 & 43.47~$\pm$~2.94\\
    & Co-teaching & 53.81~$\pm$~0.76 & 51.22~$\pm$~0.61 & 53.90~$\pm$~0.45 & 39.24~$\pm$~1.19 & 53.99~$\pm$~0.47 & 48.92~$\pm$~2.04\\
    & SIGUA & 51.33~$\pm$~1.41 & 47.47~$\pm$~2.35 & 50.81~$\pm$~2.19 & 32.12~$\pm$~4.37 & 51.22~$\pm$~2.61 & 30.82~$\pm$~7.75\\
    & Co-Dis & 54.20~$\pm$~0.39 & 51.97~$\pm$~0.46 & 54.30~$\pm$~0.15 & 41.04~$\pm$~1.77 & 54.25~$\pm$~0.25 & \textbf{49.03~$\pm$~1.76}\\\cmidrule{2-8}
    & RT-Catoni's & \textbf{57.83~$\pm$~0.45} & \textbf{53.16~$\pm$~0.74} & \textbf{54.95~$\pm$~0.85} & \underline{\textbf{44.25~$\pm$~2.36}} & 56.68~$\pm$~0.58 & 48.85~$\pm$~1.21\\
    & RT-Log-sum & \textbf{58.07~$\pm$~0.32} & \textbf{53.30~$\pm$~0.48} & \textbf{55.22~$\pm$~0.31} & \textbf{44.21~$\pm$~1.61} & 56.95~$\pm$~0.75 & \textbf{49.01~$\pm$~1.49}\\
    & RT-Welsch+ & \underline{\textbf{58.08~$\pm$~0.67}} & \underline{\textbf{54.22~$\pm$~0.83}} & \underline{\textbf{56.32~$\pm$~0.27}} & \textbf{42.75~$\pm$~2.35} & \underline{\textbf{57.98~$\pm$~0.57}} & \underline{\textbf{50.13~$\pm$~1.83}}\\
    \bottomrule
    \end{tabular}
    \caption{\small{Mean and standard deviations of test accuracy (\%) on synthetic \textit{MNIST}, \textit{SVHN}, \textit{CIFAR-10}, \textit{CIFAR-100}, and \textit{NEWS}. The best 3 experimental results are in bold while the best is underlined.}}
    \label{tab:synthetic}
    \vspace{-10pt}
\end{table*}

\begin{table*}[!h]
  \centering
  \scriptsize
    \begin{tabular}{l|ccc|ccc|ccc}
    \toprule
    \multirow{2}*{Methods} & \multicolumn{3}{c|}{RT-Catoni's} & \multicolumn{3}{c|}{RT-Log-sum} & \multicolumn{3}{c}{RT-Welsch+} \\
    \cmidrule{2-10}
     ~ & Sym.-30\% & Pair.-30\% & Ins.-30\% & Sym.-30\% & Pair.-30\% & Ins.-30\% & Sym.-30\% & Pair.-30\% & Ins.-30\%\\
    \midrule
    Gaussian & 89.36~$\pm$~0.04 & 90.77~$\pm$~0.27 & 89.25~$\pm$~0.37 & 89.42~$\pm$0.10 & 90.65~$\pm$~0.21 & 89.66~$\pm$~0.17 & 90.38~$\pm$~0.34 & 89.19~$\pm$~0.09 & 88.77~$\pm$~0.42\\
     \midrule
    Fixed & 89.05~$\pm$~0.14 & 90.83~$\pm$~0.20 & 89.07~$\pm$~0.77 & 89.60~$\pm$~0.44 & 90.49~$\pm$~0.12 & 89.65~$\pm$~0.88 & 90.65~$\pm$~0.22 & 90.15~$\pm$~0.38 & 89.56~$\pm$~0.62
    \\
    \bottomrule
    \end{tabular}%
  \label{tab:stable}%
  \vspace{-5pt}
  \caption{\small{Mean and standard deviations of test accuracy (\%) with different parameter determination ways.}}
  \vspace{-5pt}
  \label{tab:stable}%
\end{table*}%

\begin{table*}[!h]
  \centering
  \scriptsize
    \begin{tabular}{l|ccc|ccc|ccc}
    \toprule
    \multirow{2}*{Methods} & \multicolumn{3}{c}{Catoni's Based} & \multicolumn{3}{|c|}{Log-sum Based} & \multicolumn{3}{c}{Welsch+ Based} \\
    \cmidrule{2-10}
     ~ & Sym.-30\% & Pair.-30\% & Ins.-30\% & Sym.-30\% & Pair.-30\% & Ins.-30\% & Sym.-30\% & Pair.-30\% & Ins.-30\%\\
    \midrule
    Original & 84.95~$\pm$~0.59 & 85.37~$\pm$~0.78 & 84.25~$\pm$~0.66 & 85.31~$\pm$~0.56 & 54.64~$\pm$~0.58 & 84.19~$\pm$~0.86 & 85.51~$\pm$~0.40 & 71.91~$\pm$~5.79 & 50.85~$\pm$~18.60\\
    \midrule
    T-CE & 85.07~$\pm$~0.31 & 86.32~$\pm$~0.19 & 86.15~$\pm$~1.48 & 85.77~$\pm$~0.35 & 82.30~$\pm$~1.95 & 86.33~$\pm$~1.37 & 85.55~$\pm$~0.89 & 84.12~$\pm$~2.93 & 70.50~$\pm$~4.05
    \\
   \midrule
    T-M-estimators & 86.43~$\pm$~0.35 & 90.25~$\pm$~0.22 & 88.37~$\pm$~1.71 & 86.60~$\pm$~0.41 & 88.44~$\pm$~1.45 & 88.43~$\pm$~1.72 & 86.74~$\pm$~0.33 & 88.41~$\pm$~1.48 & 88.50~$\pm$~1.53
    \\
    \midrule
    RT-M-estimators & 89.39~$\pm$~0.28 & 90.83~$\pm$~0.20 & 89.34~$\pm$~0.32 & 89.60~$\pm$~0.44 & 90.49~$\pm$~0.12 & 89.65~$\pm$~0.88 & 90.65~$\pm$~0.22 & 90.15~$\pm$~0.38 & 89.56~$\pm$~0.62
    \\
    \bottomrule
    \end{tabular}%
  \label{tab:ablation}%
  \vspace{-5pt}
  \caption{\small{Mean and standard deviations of test accuracy (\%) with M-estimators (i.e., ``Original''), truncated CE (abbreviated as ``T-CE''), truncated M-estimators (abbreviated as ``T-M-estimators''), and regularly truncated M-estimators (abbreviated as ``RT-M-estimators'').} The experiments are conducted on synthetic \textit{CIFAR-10}.}
  \vspace{-5pt}
  \label{tab:ablation}%
\end{table*}%

\noindent\textbf{Generating noisy labels.} We consider two kinds of \textit{class-dependent} label noise and one kind of \textit{instance-dependent} label noise here. (1) Symmetric noise (abbreviated as Sym.) \cite{patrini2017making}: this kind of label noise is generated by flipping labels in each class uniformly to incorrect labels of other classes. (2) Pairflip noise (abbreviated as Pair.) \cite{han2018co,yu2019does}: the noise flips each class to its adjacent class. (3) Instance noise (abbreviated as Ins.) \cite{cheng2017learning}: the noise is quite realistic, where the probability that an instance is mislabeled depends on its instances/features. We generate this type of label noise as did in \cite{xia2020part}. For symmetric noise and instance noise, we set the noise rate $\tau$ to 30\% and 50\%. While, for pairflip noise, we set the noise rate $\tau$ to 30\% and 45\%, which aims to ensure that clean labels are diagonally dominant in noisy classes \cite{han2018co,xia2021sample,wu2020class2simi}. We leave out 10\% of the noisy training data a validation set, which is used for model selection. Note that the correct labels are dominating in each noisy class and that label noise is random, the accuracy of the noisy validation set and the accuracy of the clean test data set are positively correlated. The noisy validation set therefore can be employed in experiments.

\noindent\textbf{Network structure and optimizer.} In terms of the five datasets with synthetic noise, for \textit{MNIST}, we use a 3-layer MLP. Following \cite{xia2021robust}, for \textit{SVHN} and \textit{CIFAR-10}, a ResNet-18 network is used. For \textit{CIFAR-100}, a ResNet-50 network is used. Also, we employ a 3-layer MLP with the Softsign active function as did in \cite{wu2020class2simi}. We use SGD with momentum 0.9, weight decay $10^{-3}$, batch size 128, and an initial learning rate $10^{-2}$ to train the networks. The learning rate is divided by 10 after the 40th epochs and 80th epochs. The maximum number of epochs is set to 200. For \textit{SVHN}, \textit{CIFAR-10}, and \textit{CIFAR-100}, we perform data augmentation by horizontal random flips and 32$\times$32 random crops after padding with 4 pixels on each side.

\noindent\textbf{Measurement.} As for performance measurement, we use test accuracy, i.e., \textit{test accuracy = (\# of correct prediction) / (\# of testing)}. All experiments are repeated five times. Intuitively, higher test accuracy means that the algorithm is more robust to noisy labels. We report the mean and standard deviation of the results. Besides, for fair comparison, we implement all methods with default parameters by PyTorch, and conduct all the experiments on NVIDIA Tesla V100 GPUs.

\subsubsection{Analyses of experimental results}
The results on \textit{MNIST}, \textit{SVHN}, \textit{CIFAR-10}, \textit{CIFAR-100}, and \textit{NEWS} are presented in Table~\ref{tab:synthetic}. For \textit{MNIST}, as can be seen, the proposed methods achieve competitive classification performance. For \textit{SVHN}, our methods outperform all baselines in all cases (one of the proposed methods works the best), which shows the effectiveness of our methods. For \textit{CIFAR-10} and \textit{CIFAR-100}, our methods also perform best. Lastly, for \textit{NEWS}, our methods achieve the best results. Almost all the experimental results justify our claims well. Note that the performance on \textit{NEWS} is a bit different from the results in~\cite{xia2023co}. This is because the optimization of the two works is different. We use the SGD optimizer with momentum, while \cite{xia2023co} uses the Adam optimizer.

\begin{figure}[!t]
\centering
\begin{subfigure}{.465\linewidth}\label{fig:selection_a}
  \centering
  \includegraphics[width=1\linewidth]{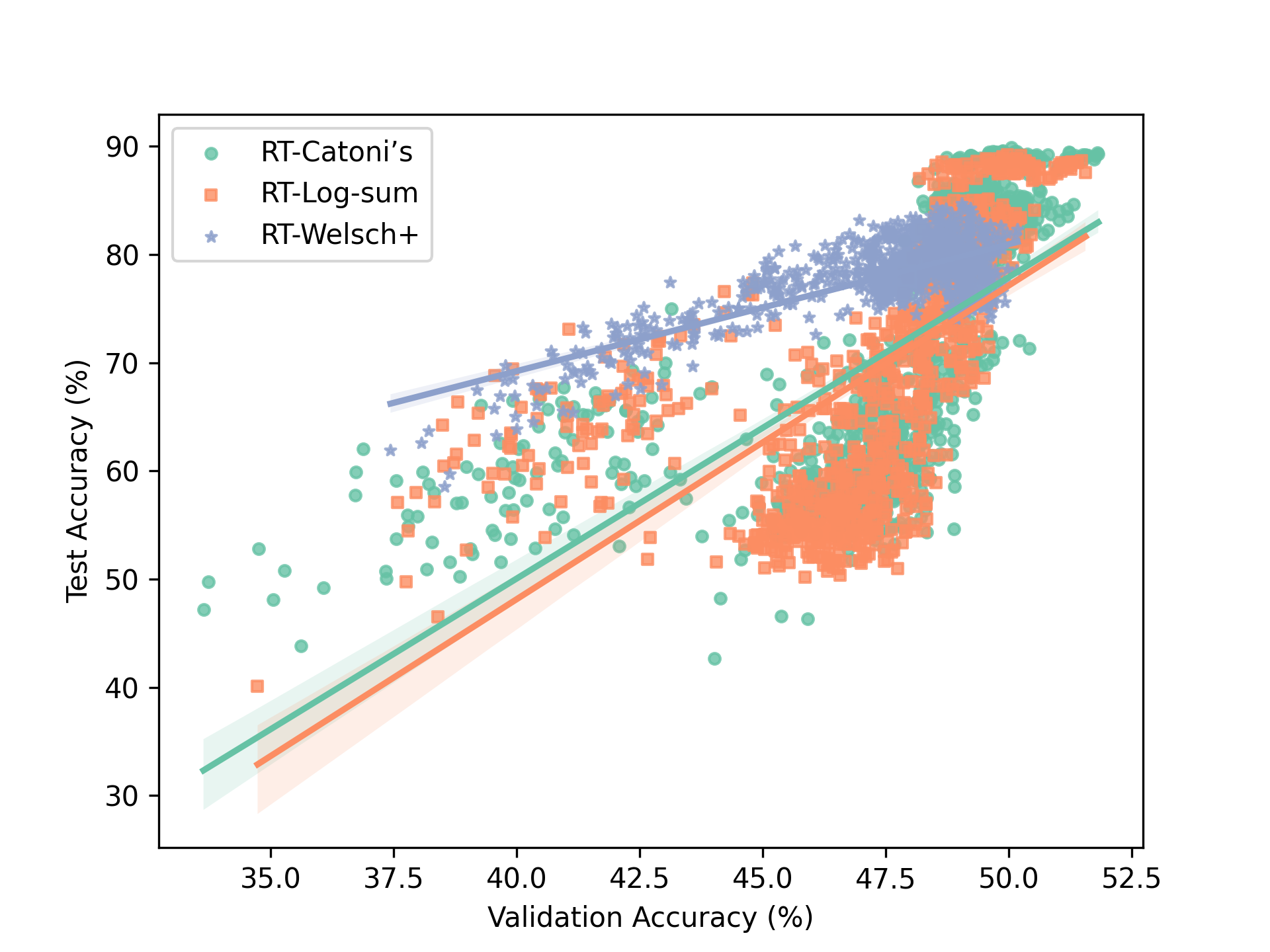} 
  \caption{}
\end{subfigure}
\hspace{4mm}
\begin{subfigure}{.465\linewidth}\label{fig:selection_b}
  \centering
  \includegraphics[width=1\linewidth]{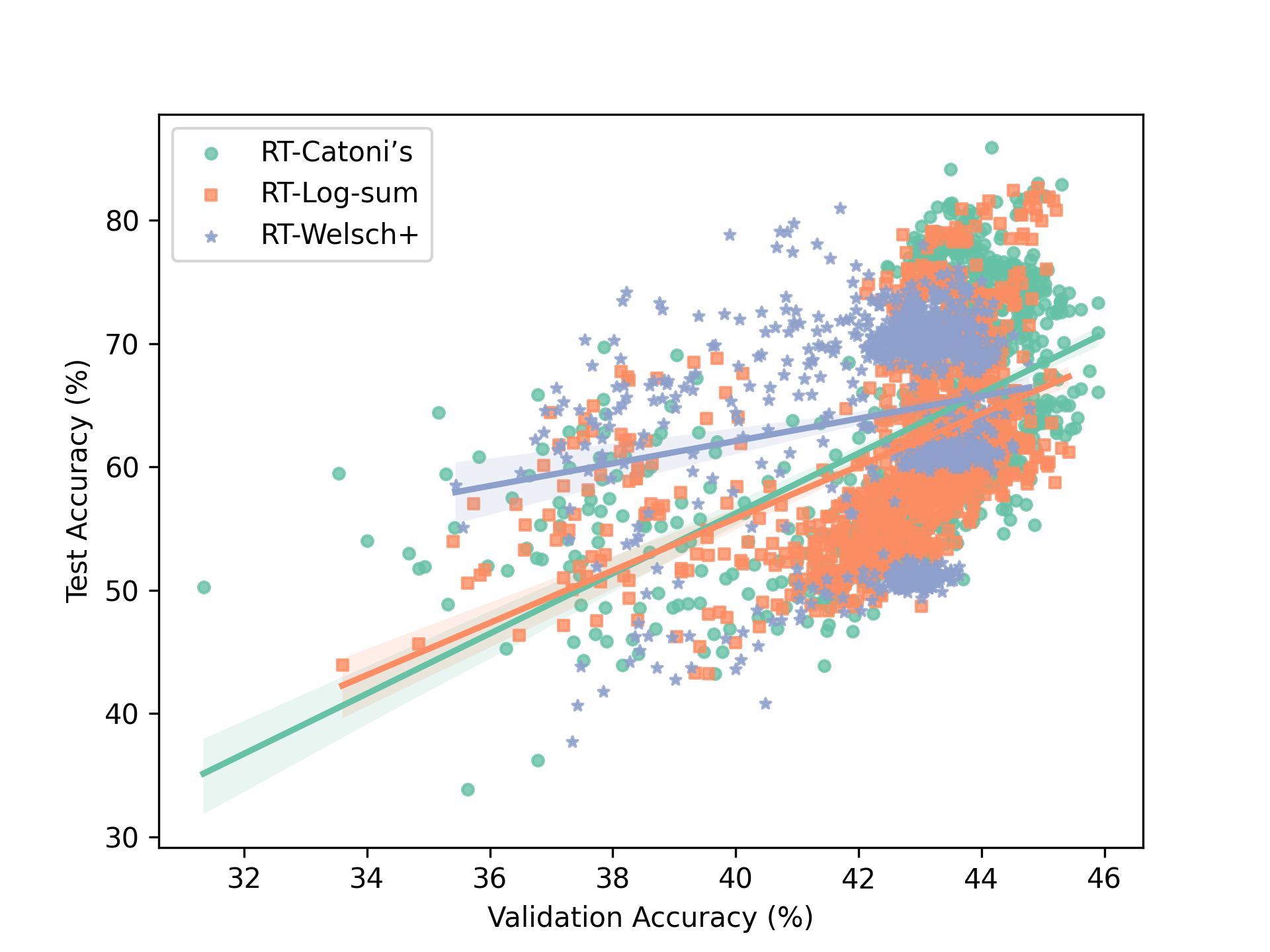} 
  \caption{}
\end{subfigure}
\vspace{-5pt}
\caption{\small{Illustrations of positive correlation between validation accuracy and test accuracy. Straight lines are achieved by regression, which reflect the overall trend. The experiments are conducted on synthetic \textit{CIFAR-10} with Pair.-45\% noise~(subfigure~(a)) and Ins.-50\% noise (subfigure~(b)).}}
\vspace{-7pt}
\label{fig:selection}
\end{figure}

\subsubsection{Discussions of method selection}
Note that different methods built on different M-estimators perform variably in different label noise cases. As discussed before, the differences between different M-estimators lie in the different ranges with respect to the training loss and different punishments on larger-loss examples. If we tend to choose the most suitable M-estimator, we need prior knowledge of data distributions, network architectures, and training dynamics (e.g., the loss distribution during optimization), which is rather hard or even impossible in practice. Fortunately, all methods built on M-estimators exhibit superior performance over baselines in most cases, which demonstrates the effectiveness of our M-estimator-based framework. 

Moreover, here we propose to use the accuracy achieved on the noisy validation set for the selection of different methods. It is because the accuracy of the noisy validation set and the accuracy of
the clean test data set are positively correlated. Therefore, for three methods, with the same training and validation data, we can choose the method that \textit{overall} enjoys the higher validation accuracy. In Fig.~\ref{fig:selection}, using the regression technology to mitigate the randomness and reflect the overall trend, we show the positive correlation between the accuracies of the noisy validation and test sets for method selection. Also, our RT-Catoni's enjoys both the highest validation accuracy and the highest test accuracy, which matches the results in Table~\ref{tab:synthetic}.

\subsubsection{The stability of our method}
\noindent\textbf{The stability about $\sigma$.} As discussed in Section \ref{sec:3.2}, our methods keep an adaptive manner to perform truncation and sample selection, i.e., using the ``three-sigma'' rule to determine a threshold. As an adaptive method, we do not need to estimate the noise rate. Prior works on sample selection show that if the noise rate cannot be estimated accurately, the classification performance will be affected largely \cite{yao2020searching}. Here, we show that our methods are stable even though the threshold is changed artificially during training. 

\begin{figure}[!h]
\centering
\begin{subfigure}{\linewidth}\label{fig:}
  \centering
  \includegraphics[width=1\linewidth]{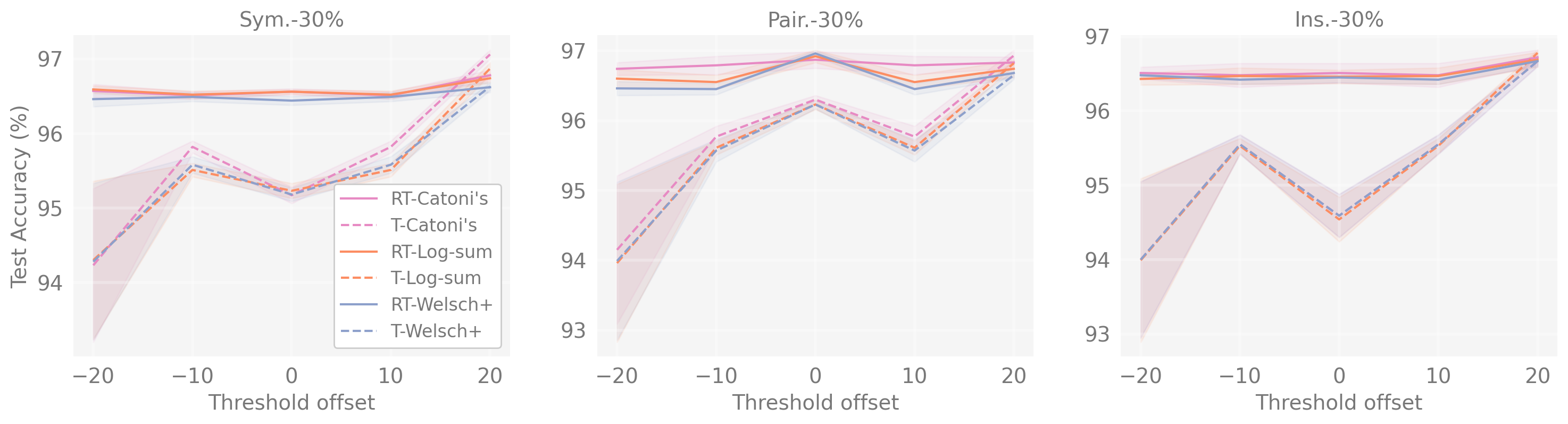} 
  \caption{}
\end{subfigure}
\begin{subfigure}{\linewidth}\label{fig:}
  \centering
  \includegraphics[width=1\linewidth]{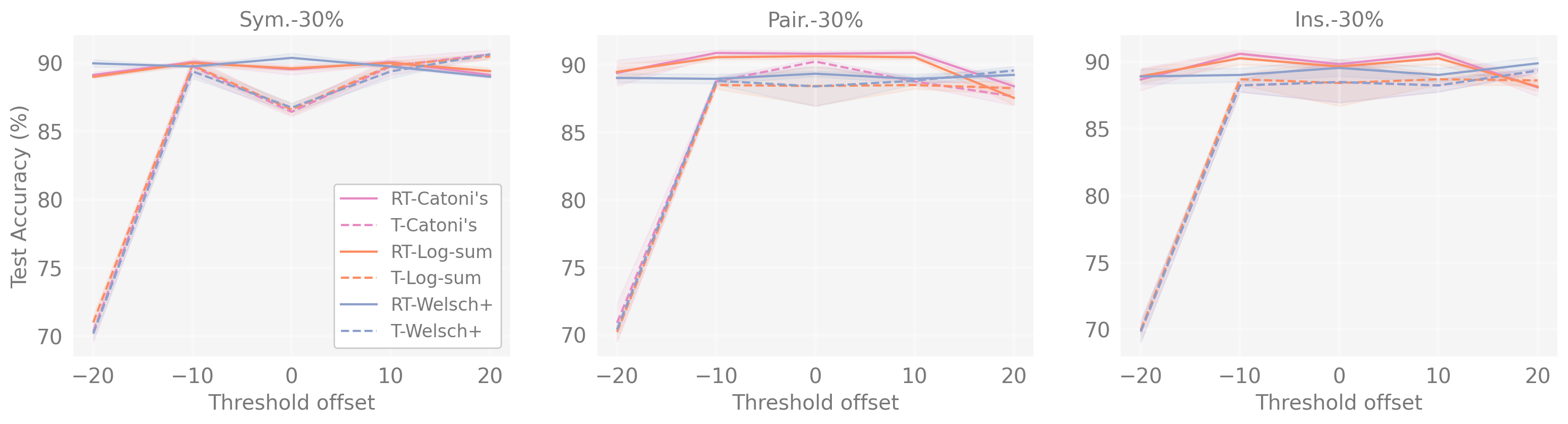} 
  \caption{}
\end{subfigure}
\caption{\small{Illustrations of the test accuracy with different disturbances. The experimental results reveal that regularly truncated M-estimators are more stable. The experiments are conducted on synthetic \textit{MNIST} (subfigure (a)) and synthetic \textit{CIFAR-10} (subfigure (b)) with 30\% noise rates.}}
\label{fig:stable}
\end{figure}

The experiments are conducted on \textit{MNIST} and \textit{CIFAR-10} with 30\% noise rates. Let $\Delta_\sigma$ be the disturbance added to the $\sigma$, where $\sigma$ denotes the threshold determined by the algorithms. As shown in Fig.~\ref{fig:stable}, the truncated M-estimators are very sensitive to the values of thresholds. In particular, when $\Delta_\sigma=-20\%$, the classification performance of truncated M-estimators is greatly affected. On synthetic \textit{CIFAR-10}, the test accuracies are reduced by almost 20\%. As a comparison, the proposed regularly truncated M-estimators are very stable when the disturbance is added to the threshold. It is because we regularly introduce large-loss examples into training. The underlying clean examples can be exploited. Also, such a way can address the covariate shift issue effectively mentioned in \cite{jiang2018mentornet}, and therefore helps generalization.  

\noindent\textbf{The stability about $\epsilon$ and $\alpha$.} We present the sensitivity analyses on the intrinsic parameters of exploited M-estimators, i.e., $\epsilon$ in Log-sum Penalty and $\alpha$ in Welsch+. The experiments are conducted on \textit{MNIST} and \textit{CIFAR-10} with 30\% noise rates. The range of $\epsilon$ and $\alpha$ is \{1.5,2,2.5,3\}. As can be seen in Tables~\ref{tab:sensitivity1} and \ref{tab:sensitivity2}, the M-estimators are robust to the choice of intrinsic parameters in a certain range, which implies that the proposed methods can be easily applied in practice.

\begin{table*}[!t]
    \centering
    \tiny
    \renewcommand{\arraystretch}{0.9}
    \begin{tabular}{l|cccc|cccc|cccc}
    \toprule
        Methods & \multicolumn{4}{c|}{Sym.-30\%} & \multicolumn{4}{c|}{Pair.-30\%} & \multicolumn{4}{c}{Ins.-30\%} \\\midrule
        \multirow{2}{*}{RT-Log-sum} &$\epsilon=1.5$&$\epsilon=2$&$\epsilon=2.5$&$\epsilon=3$&$\epsilon=1.5$&$\epsilon=2$&$\epsilon=2.5$&$\epsilon=3$&$\epsilon=1.5$&$\epsilon=2$&$\epsilon=2.5$&$\epsilon=3$\\\cmidrule{2-13}
        &96.54 $\pm$ 0.35&96.47 $\pm$ 0.40&96.49 $\pm$ 0.40&96.52 $\pm$ 0.38&96.50 $\pm$ 0.23&96.40 $\pm$ 0.25&96.35 $\pm$ 0.24&96.31 $\pm$ 0.25&96.26 $\pm$ 0.32&96.22 $\pm$ 0.36&96.17 $\pm$ 0.32&96.14 $\pm$ 0.41\\\midrule
        \multirow{2}{*}{RT-Welsch+} & $\alpha=1.5$&$\alpha=2$&$\alpha=2.5$&$\alpha=3$&$\alpha=1.5$&$\alpha=2$&$\alpha=2.5$&$\alpha=3$&$\alpha=1.5$&$\alpha=2$&$\alpha=2.5$&$\alpha=3$\\\cmidrule{2-13}
        &96.55~$\pm$~0.19&96.68~$\pm$~0.33&96.71~$\pm$~0.29&96.72~$\pm$~0.33&96.52~$\pm$~0.27&96.50~$\pm$~0.24&96.32~$\pm$~0.18&96.31~$\pm$~0.30&96.42~$\pm$~0.33&96.41~$\pm$~0.33&96.35~$\pm$~0.36&96.35~$\pm$~0.42\\\bottomrule
    \end{tabular}
    \vspace{-5pt}
    \caption{\small{The sensitivity analyses on the intrinsic parameters of exploited M-estimators. The experiments are conducted on synthetic \textit{MNIST} with 30\% noise rates.}}
    \label{tab:sensitivity1}
    \vspace{-5pt}
\end{table*}

\begin{table*}[!t]
    \centering
    \tiny
    \renewcommand{\arraystretch}{0.9}
    \begin{tabular}{l|cccc|cccc|cccc}
    \toprule
        Methods & \multicolumn{4}{c|}{Sym.-30\%} & \multicolumn{4}{c|}{Pair.-30\%} & \multicolumn{4}{c}{Ins.-30\%} \\\midrule
        \multirow{2}*{RT-Log-sum} &$\epsilon=1.5$&$\epsilon=2$&$\epsilon=2.5$&$\epsilon=3$&$\epsilon=1.5$&$\epsilon=2$&$\epsilon=2.5$&$\epsilon=3$&$\epsilon=1.5$&$\epsilon=2$&$\epsilon=2.5$&$\epsilon=3$\\\cmidrule{2-13}
        &89.46~$\pm$~0.11&88.93~$\pm$~0.19&88.94~$\pm$~0.21&88.94~$\pm$~0.09&90.73~$\pm$~0.15&90.70~$\pm$~0.15&90.66~$\pm$~0.10&90.71~$\pm$~0.21&89.35~$\pm$~0.39&89.39~$\pm$~0.49&88.86~$\pm$~0.33&88.96~$\pm$~0.49\\\midrule
        \multirow{2}*{RT-Welsch+} & $\alpha=1.5$&$\alpha=2$&$\alpha=2.5$&$\alpha=3$&$\alpha=1.5$&$\alpha=2$&$\alpha=2.5$&$\alpha=3$&$\alpha=1.5$&$\alpha=2$&$\alpha=2.5$&$\alpha=3$\\\cmidrule{2-13}
        &90.81~$\pm$~0.15&90.64~$\pm$~0.11&90.15~$\pm$~0.14&89.77~$\pm$~0.19&89.76~$\pm$~0.19&90.46~$\pm$~0.20&90.40~$\pm$~0.06&90.61~$\pm$~0.28&89.48~$\pm$~0.41&89.44~$\pm$~0.72&89.63~$\pm$~0.36&89.53~$\pm$~0.21\\\bottomrule
    \end{tabular}
    \vspace{-5pt}
    \caption{\small{The sensitivity analyses on the intrinsic parameters of exploited M-estimators. The experiments are conducted on synthetic \textit{CIFAR-10} with 30\% noise rates.}}
    \label{tab:sensitivity2}
    \vspace{-5pt}
\end{table*}

\subsubsection{Ablation study}
We conduct detailed ablation studies to analyze and show the effects of different components to provide insights into what makes our methods successful. 

\begin{figure}[!h]
\centering
\begin{subfigure}{\linewidth}\label{fig:}
  \centering
  \includegraphics[width=1\linewidth]{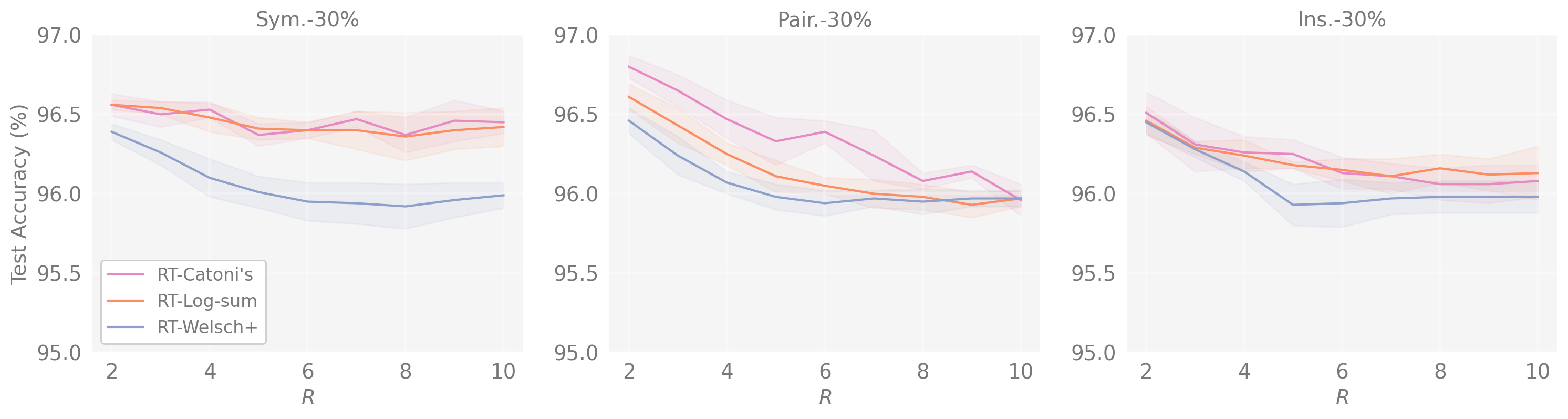} 
  \caption{}
\end{subfigure}
\begin{subfigure}{\linewidth}\label{fig:}
  \centering
  \includegraphics[width=1\linewidth]{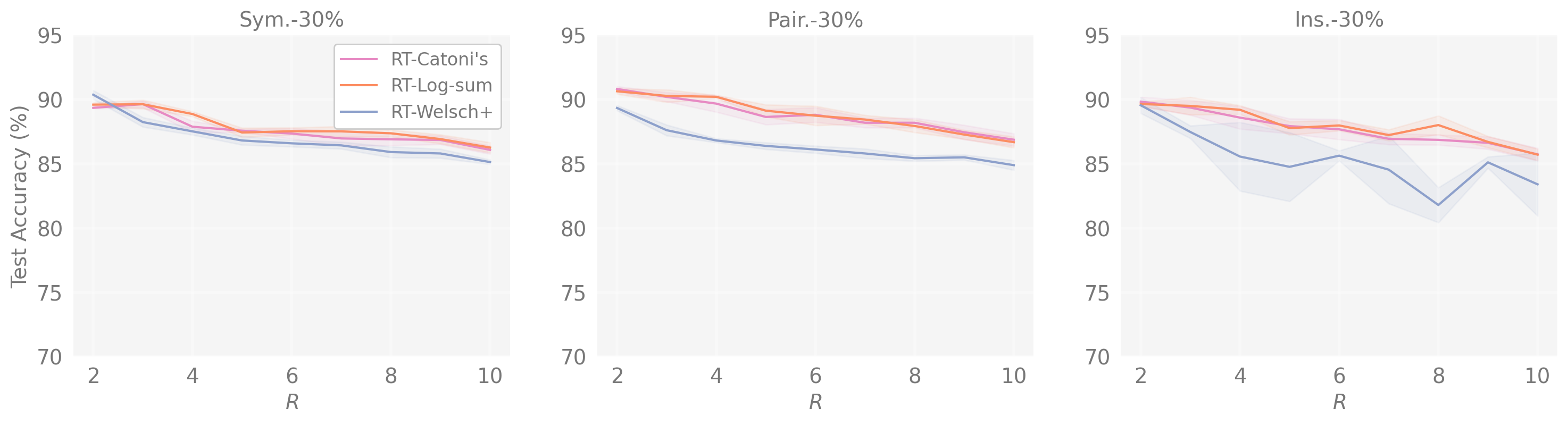} 
  \caption{}
\end{subfigure}
\caption{\small{Illustrations of the test accuracy with different values of $R$. These experiments reveal a smaller $R$, which means that introducing large-loss examples frequently can lead to better classification performance in general. The experiments are conducted on synthetic \textit{MNIST} (subfigure (a)) and synthetic \textit{CIFAR-10} (subfigure (b)) with 30\% noise rates.}}
\label{fig:ablation}
\end{figure}

\noindent\textbf{The influence of $R$.} We first analyze the effect of the frequency of using different kinds of robust M-estimators, i.e., $R$. The experimental results are shown in Fig.~\ref{fig:ablation}. As can be seen, with the increase of $R$, the test accuracies decrease clearly. In other words, the introduction of large-loss examples in a conservative way can improve the algorithm performance, which verifies the effectiveness of our methods.

\noindent\textbf{Impact of each component.} We then compare the results of M-estimators, truncated CE, truncated M-estimators, and the proposed regularly truncated M-estimators. The results are shown in Table \ref{tab:stable}. We can see that original M-estimators cannot work well when there are noisy labels, and truncated M-estimators can better handle noisy labels. Also, comparing the truncated CE with truncated M-estimators, we can see that assigning different weights on small-loss examples can improve performance. Additionally, the proposed regularly truncated M-estimators outperform truncated M-estimators, which shows the effectiveness of introducing large-loss examples into training. Note that compared with the results in Fig.~\ref{fig:ablation} and  Table~\ref{tab:stable}, we can know that both the truncation and the introduction of large-loss examples are of importance against noisy labels. Besides, there is a trade-off switching frequency with a relatively small value.

\noindent\textbf{Adaptive determination of $\epsilon$ and $\alpha$.} As discussed in Section \ref{sec:3.2}, we can adaptively determine the $\epsilon$ and $\alpha$ by introducing a Gaussian distribution assumption. We use the $\ell_2$ distance to measure the distribution $\Gamma'$ to $\mathcal{N}(\mu', \delta'^2)$ in this paper. The results are provided in Table \ref{tab:ablation}. Accordingly, we can know that our methods are able to avoid tuning the hyperparameters $\epsilon$ and $\alpha$ artificially. Instead, they can be determined automatically by using the Gaussian distribution assumption. Also, our methods work well in such a way.

\subsubsection{A closer look on the memorization effect}
\begin{figure}[!h]
\centering
\begin{subfigure}{\linewidth}\label{fig}
  \centering
  \includegraphics[width=.6\linewidth]{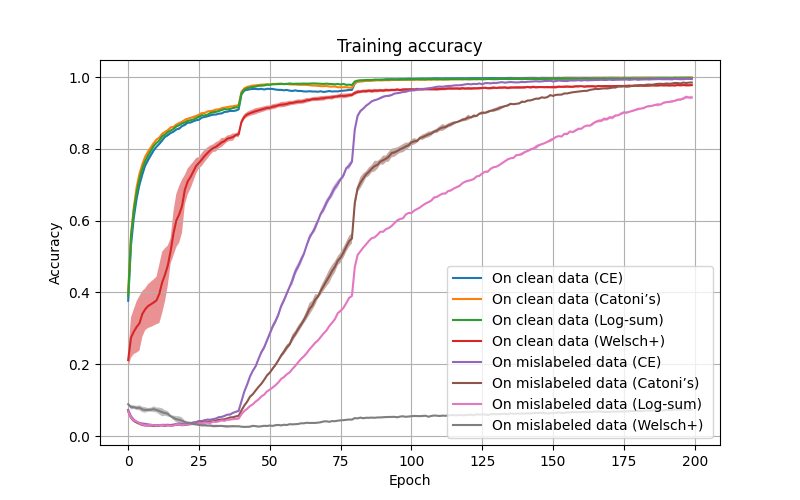} 
  \caption{}
\end{subfigure}
\begin{subfigure}{\linewidth}\label{fig:}
  \centering
  \includegraphics[width=.6\linewidth]{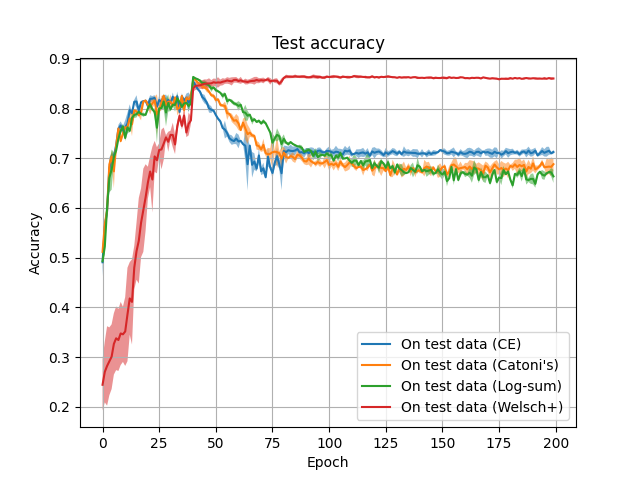} 
  \caption{}
\end{subfigure}
\caption{\small{Illustrations of the training and test accuracy achieved by different methods with the increase of epochs. Experiments are conducted on synthetic \textit{CIFAR-10} with Sym.-30\% noise. \textbf{(a):} Training accuracy \textit{vs} Epoch}. \textbf{(b):} Test accuracy \textit{vs} Epoch.}
\label{fig:dynamic}
\end{figure}

The memorization effect of the deep network~\cite{arpit2017closer} shows that it would first memorize clean data and then memorize mislabeled data. Therefore, in early training, the network is relatively robust with noisy labels, i.e., more memorization of clean data and less memorization of mislabeled data, following good test accuracy on clean test data. Here, we provide a closer look and show that exploited M-estimators (i.e., original M-estimators) can strengthen the memorization effect. The results in Fig.~\ref{fig:dynamic} show that used M-estimators make the deep network less memorize mislabeled data, leading to better test accuracy in early training. Interestingly, we find that the ways of enhancing model robustness of multiple M-estimators are slightly different, though all of them can tackle noisy labels successfully. In more detail, Catoni’s and Log-sum can strengthen the memorization of clean data and reduce the memorization of mislabeled data at the same time. Differently, Welsch+ works well in largely reducing the memorization of mislabeled data.

\begin{figure}[t]
    \centering
    \begin{minipage}[c]{0.05\columnwidth}~\end{minipage}%
    \begin{minipage}[c]{0.05\columnwidth}~\end{minipage}%
    \begin{minipage}[c]{0.3\linewidth}\centering\small  \footnotesize{--RT-Catoni's--}  \end{minipage}%
    \begin{minipage}[c]{0.3\linewidth}\centering\small \footnotesize{--RT-Log-sum--}  \end{minipage}%
    \begin{minipage}[c]{0.3\linewidth}\centering\small \footnotesize{--RT-Welsch+--}  \end{minipage}\\
    
    \begin{minipage}[c]{0.05\columnwidth}~\end{minipage}%
    \begin{minipage}[c]{0.05\columnwidth}\centering\small \rotatebox[origin=c]{90}{\footnotesize{--Sym.-30\%--}} \end{minipage}%
    \begin{minipage}[c]{0.9\linewidth}
        \includegraphics[width=0.33\linewidth]{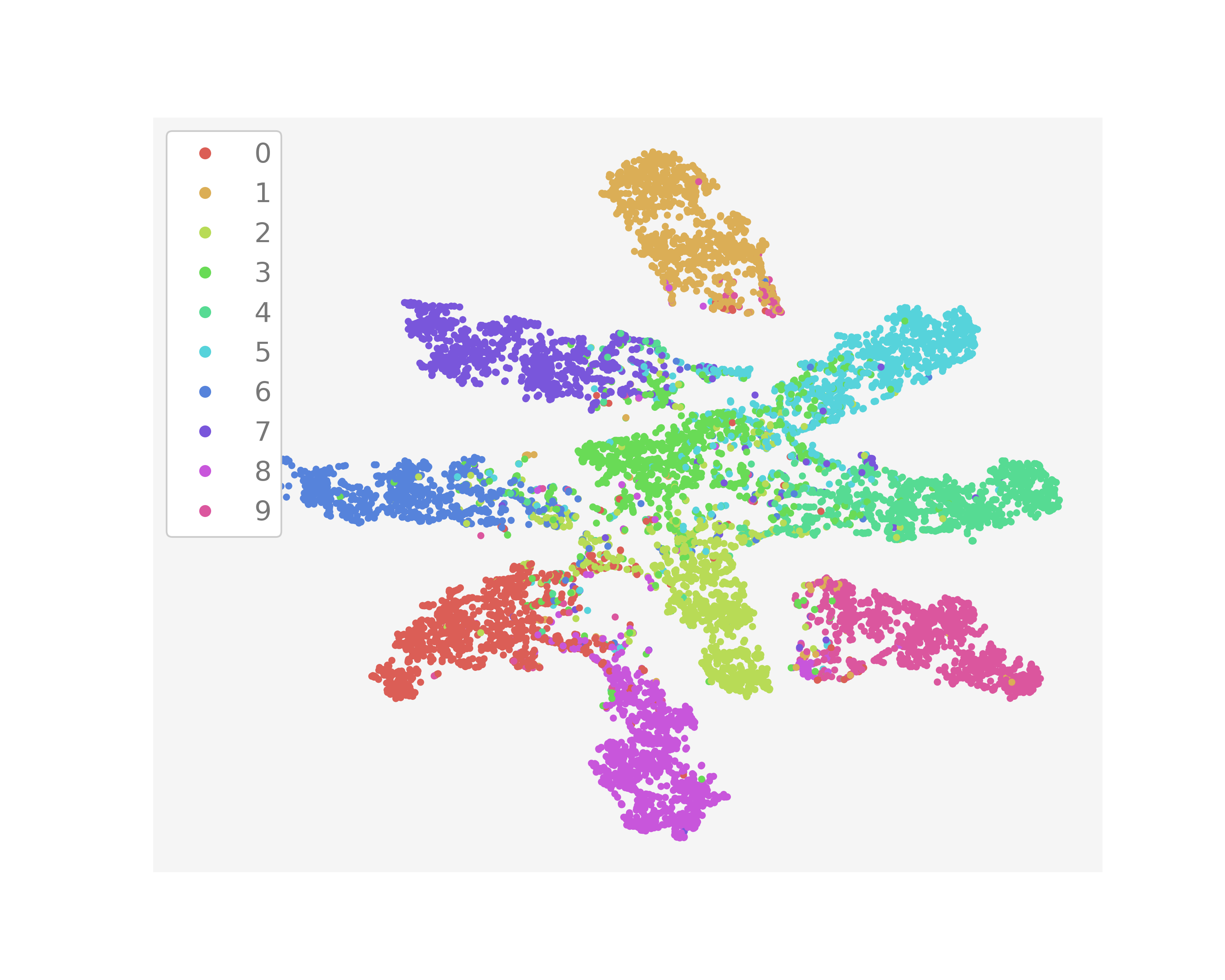}%
        \includegraphics[width=0.33\linewidth]{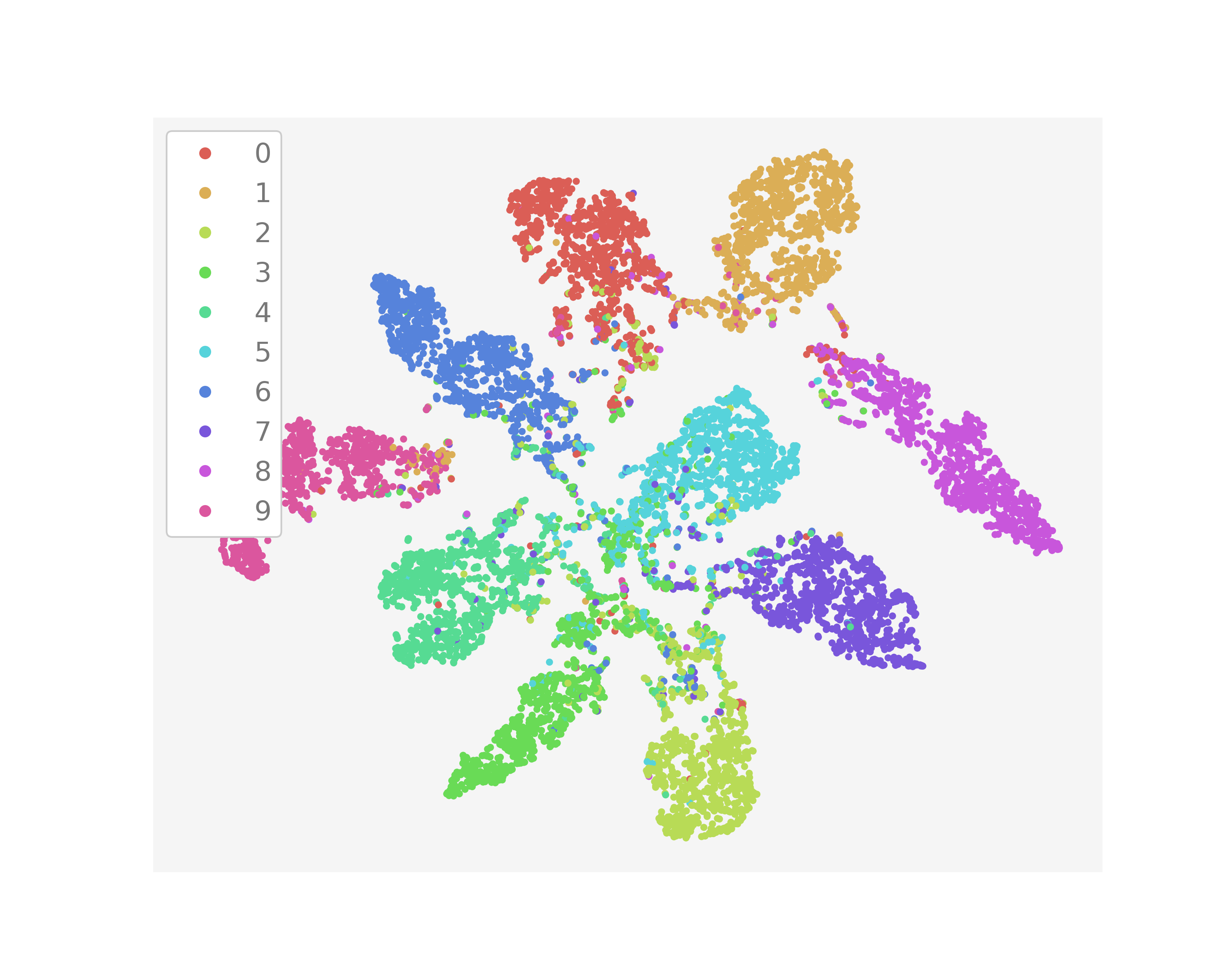}%
        \includegraphics[width=0.33\linewidth]{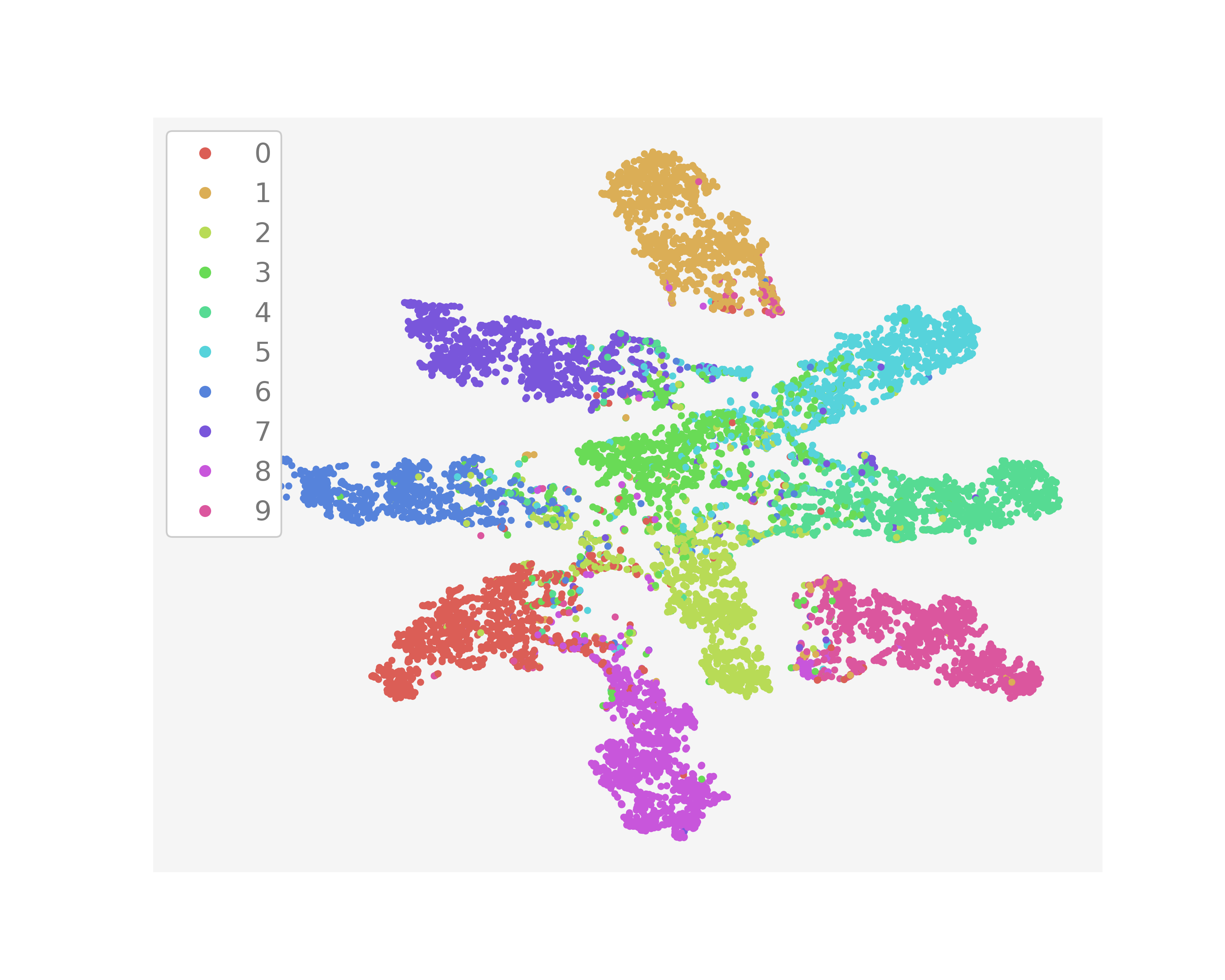}%
    \end{minipage}
    
    \begin{minipage}[c]{0.05\columnwidth}~\end{minipage}%
    \begin{minipage}[c]{0.05\columnwidth}\centering\small \rotatebox[origin=c]{90}{\footnotesize{--Pair.-30\%--}} \end{minipage}%
    \begin{minipage}[c]{0.9\linewidth}
        \includegraphics[width=0.33\linewidth]{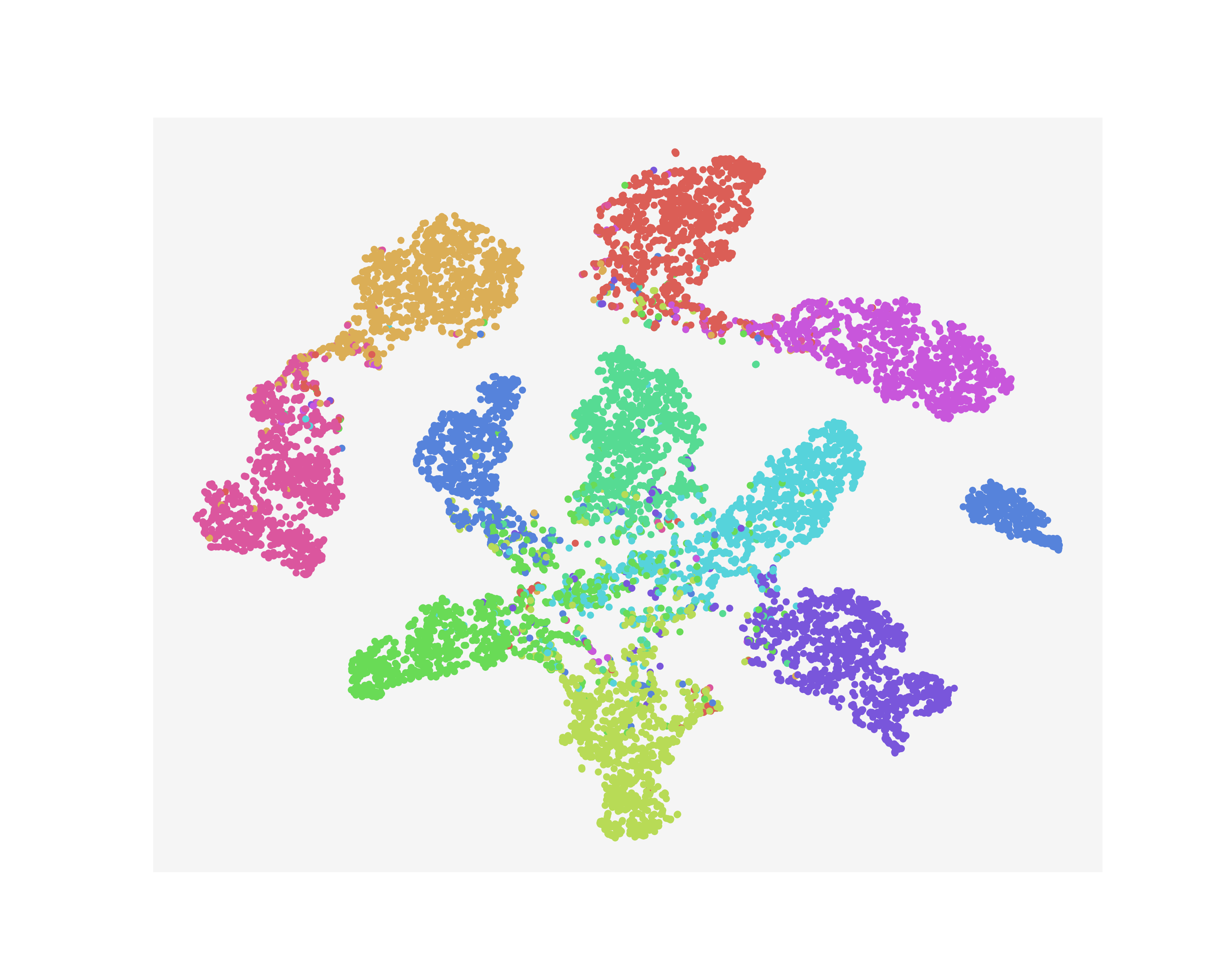}%
        \includegraphics[width=0.33\linewidth]{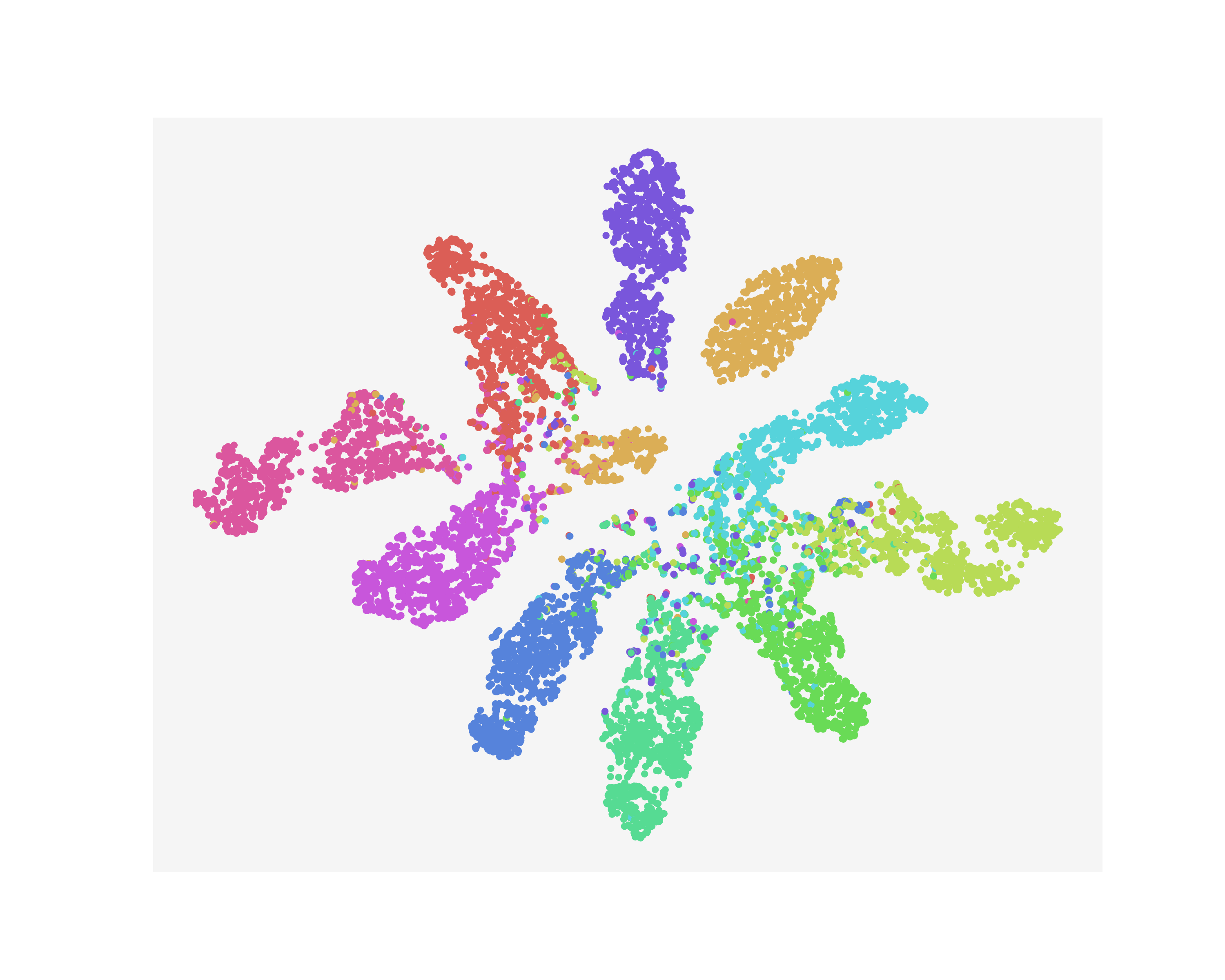}%
        \includegraphics[width=0.33\linewidth]{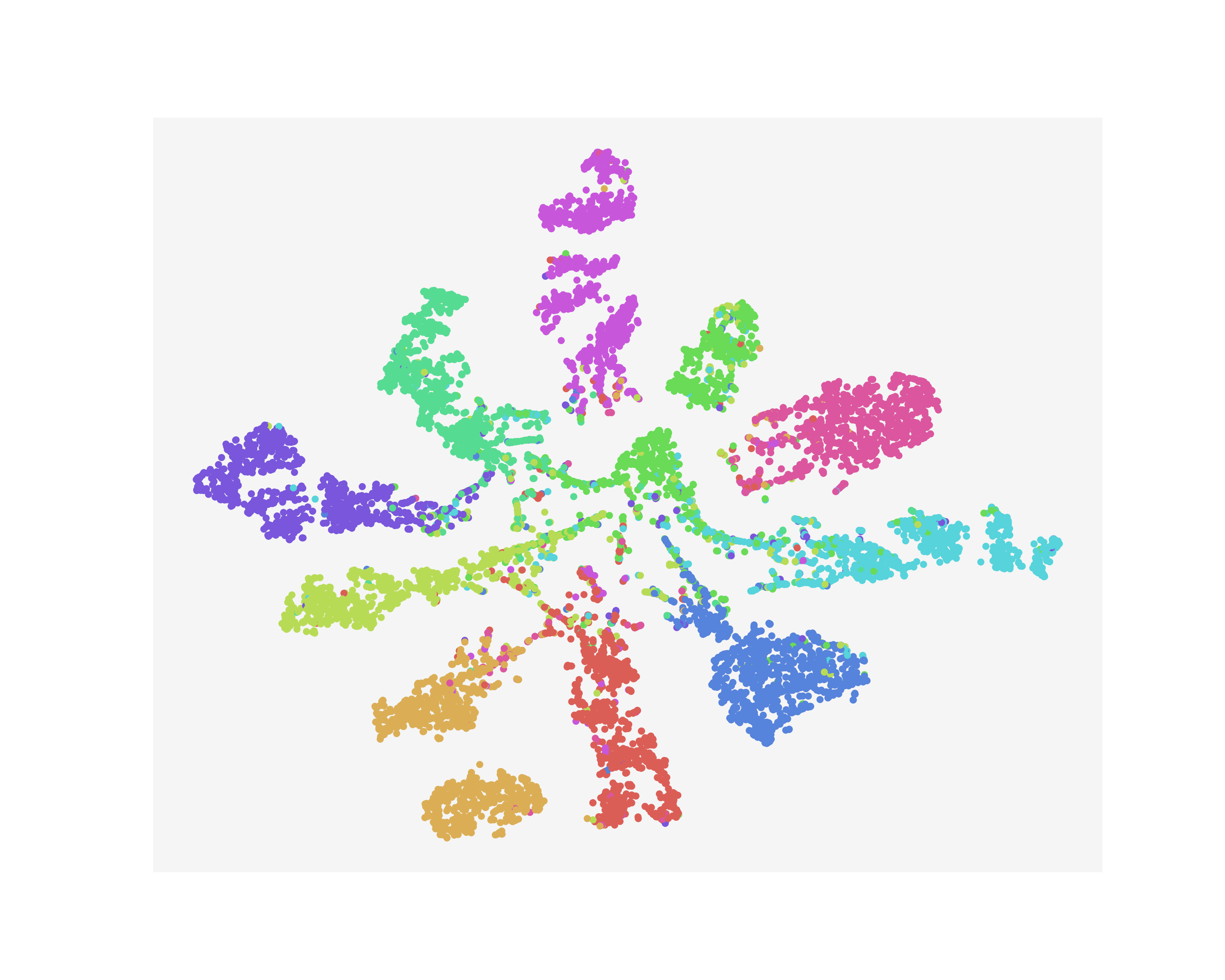}%
    \end{minipage}
    
    \begin{minipage}[c]{0.05\columnwidth}~\end{minipage}%
    \begin{minipage}[c]{0.05\columnwidth}\centering\small \rotatebox[origin=c]{90}{\footnotesize{--Ins.-30\%--}} \end{minipage}%
    \begin{minipage}[c]{0.9\linewidth}
        \includegraphics[width=0.33\linewidth]{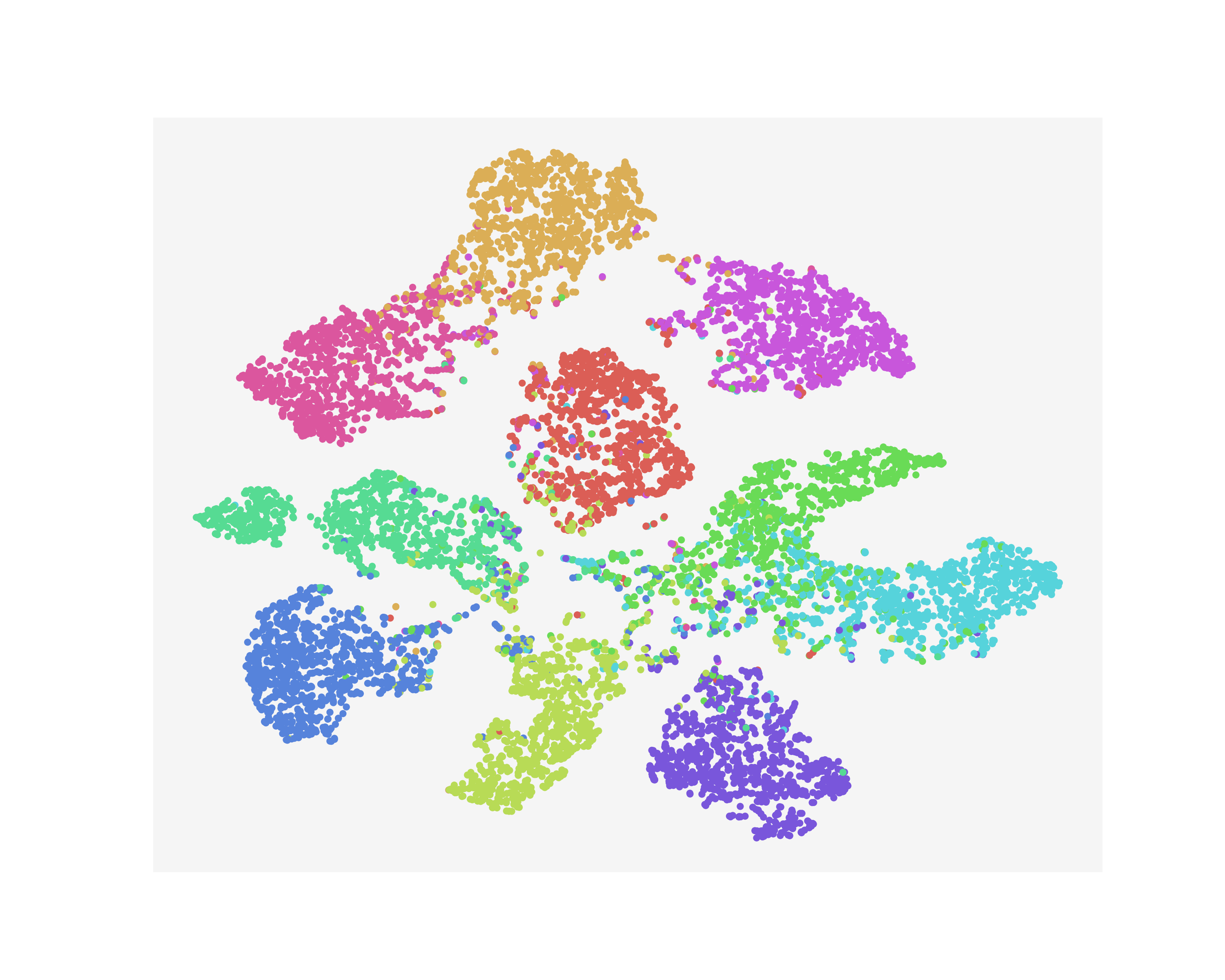}%
        \includegraphics[width=0.33\linewidth]{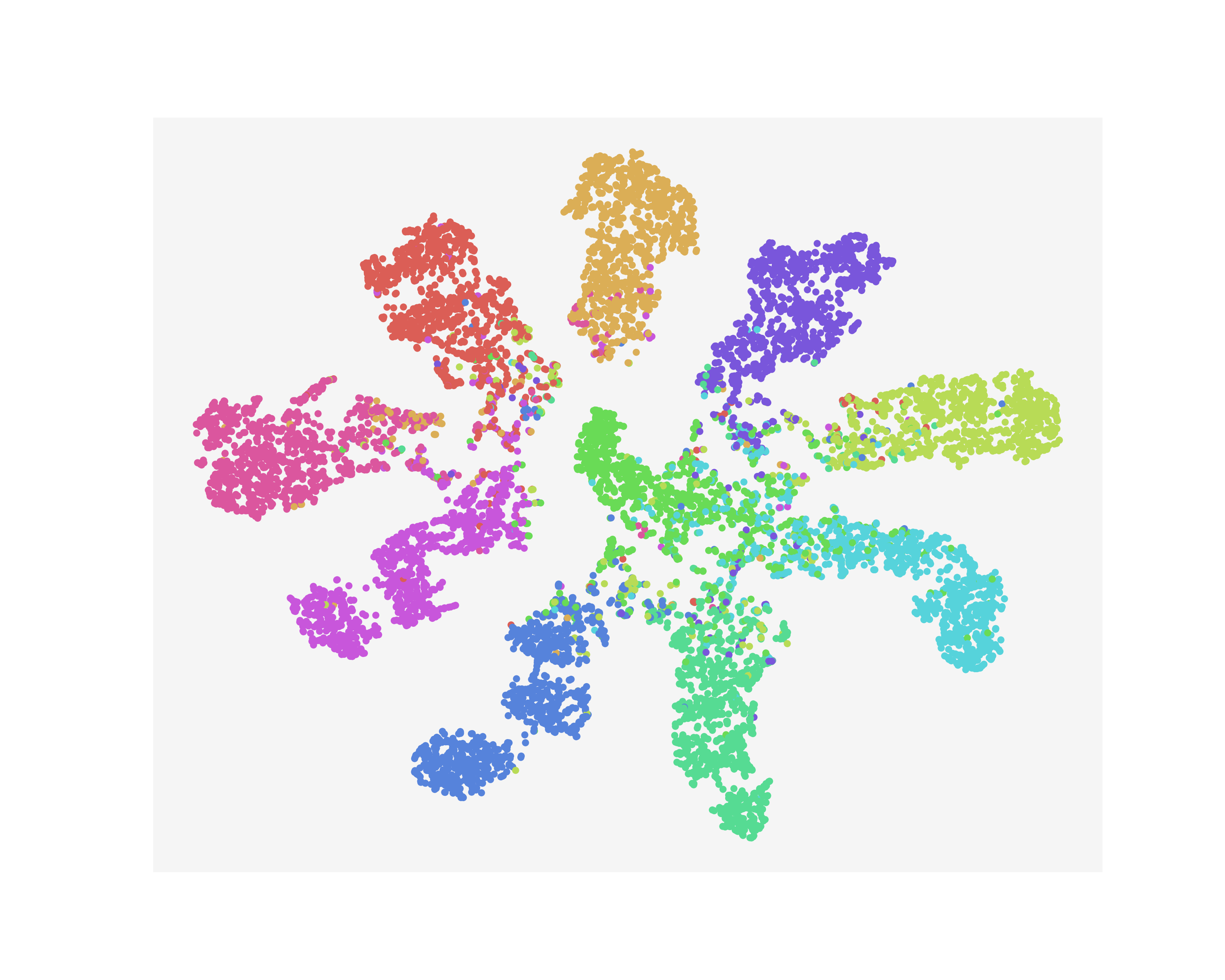}%
        \includegraphics[width=0.33\linewidth]{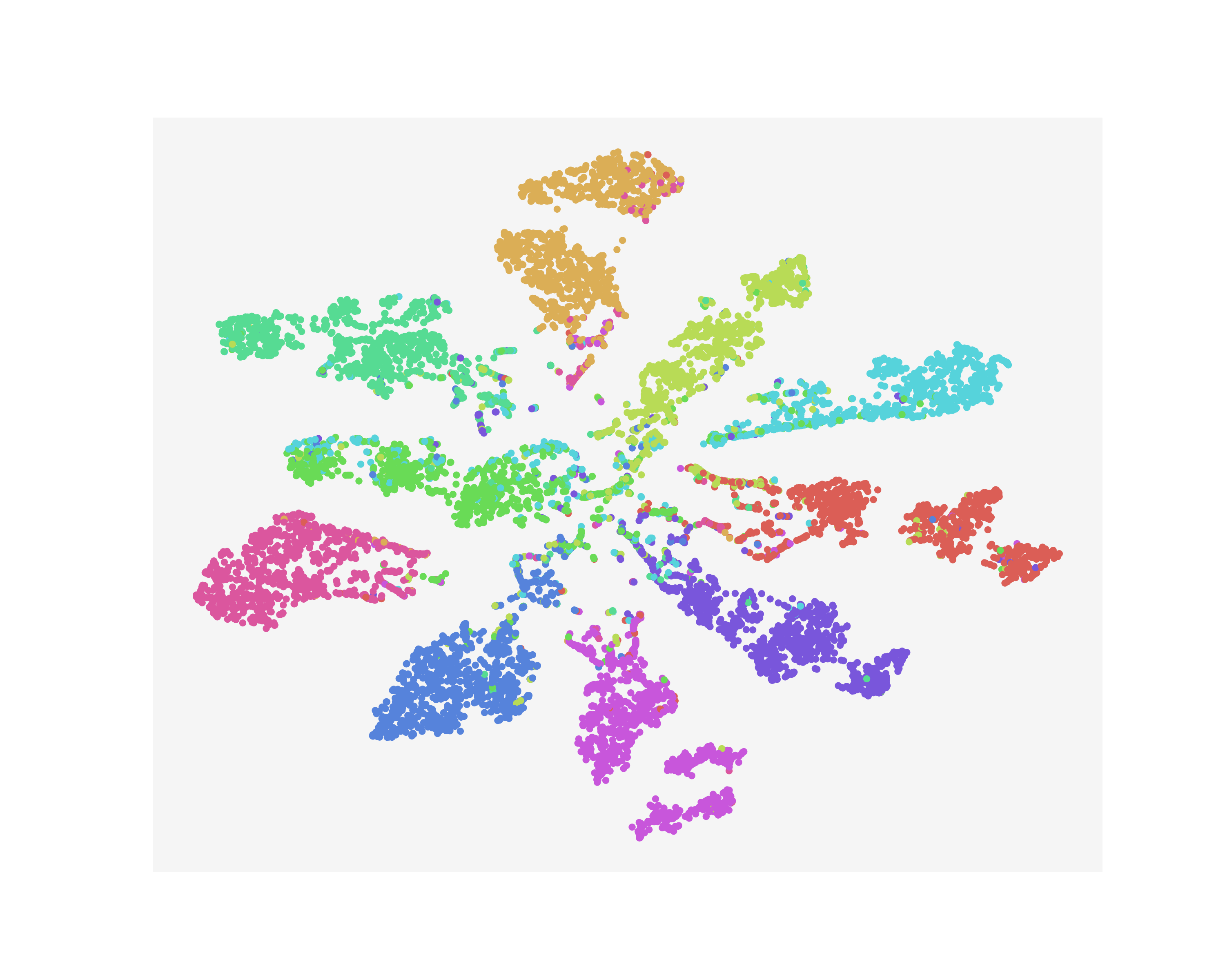}%
    \end{minipage}
    
    \caption{\small{Visualizations of experimental results using 2D t-SNE \cite{van2008visualizing}.} The experiments are conducted on synthetic \textit{CIFAR-10}.}
    \label{fig:ab}
    \vspace{-5pt}%
\end{figure}

\begin{table}[!t]
    \centering
    \scriptsize
    \begin{tabular}{l|ccc}
    \toprule
    Methods & Sym.-50\% & Pair.-45\% & Ins.-50\%\\\midrule
        DivideMix &  95.00 $\pm$ 1.12 & 86.55 $\pm$ 2.74  & \textbf{92.90  $\pm$ 2.26} \\\midrule
        DivideMix+RT-Catoni’s & \textbf{95.01 $\pm$ 1.01}  & \textbf{94.88  $\pm$ 1.63} &  \textbf{94.56  $\pm$ 2.84} \\
        DivideMix+RT-Log-sum & \underline{\textbf{95.18  $\pm$ 0.84}} & \textbf{94.84  $\pm$ 1.36} & \underline{\textbf{94.78  $\pm$ 1.29}} \\
        DivideMix+RT-Welsch & \textbf{95.02 $\pm$ 1.28} & \underline{\textbf{95.21  $\pm$ 2.77}} & 91.51 $\pm$ 2.29 \\
        \bottomrule
    \end{tabular}
    \vspace{-2pt}
    \caption{\small{Mean and standard deviations of test accuracy (\%) on \textit{CIFAR-10} compared DivideMix with the methods boosted by the proposed algorithms. The best 3 experimental results are in bold while the best is underlined.}}
    \vspace{-2pt}
    \label{tab:dividemix}
\end{table}

\begin{table}[!t]
    \centering
    \scriptsize
    \begin{tabular}{l|ccc}
    \toprule
    Methods & Sym.-50\% & Pair.-45\% & Ins.-50\%\\\midrule
        CL & 82.56 $\pm$ 1.14 & 58.97  $\pm$ 1.52 & 56.51 $\pm$ 2.82 \\\midrule
        CL+RT-Catoni’s & \textbf{87.75 $\pm$ 1.90} & \underline{\textbf{90.16  $\pm$ 2.64}} & \underline{\textbf{75.45  $\pm$ 3.05}}  \\
        CL+RT-Log-sum & \textbf{87.73 $\pm$ 1.59} & \textbf{88.96 $\pm$ 2.21} & \textbf{72.88 $\pm$ 3.15}\\
        CL+RT-Welsch & \underline{\textbf{88.26 $\pm$ 1.41}} & \textbf{80.99 $\pm$ 2.31} & \textbf{72.91 $\pm$ 4.43}\\
        \bottomrule
    \end{tabular}
    \caption{\small{Mean and standard deviations of test accuracy (\%) on \textit{CIFAR-10} compared contrastive learning~(CL) with the methods boosted by the proposed algorithms. The best 3 experimental results are in bold while the best is underlined.}}
    \vspace{-12pt}
    \label{tab:self}
\end{table}

\begin{table*}[!tp]
  \centering
  \scriptsize
    \begin{tabular}{l|c|c|cccc}
    \toprule
    Methods & \textit{Food-101} & \textit{Clothing1M} & \textit{CIFAR-10N-1} & \textit{CIFAR-10N-2} & \textit{CIFAR-10N-3} & \textit{CIFAR-10N-W} \\
    \midrule
    CE& 85.15 & 68.88 & 85.41~$\pm$~0.24 & 86.79~$\pm$~0.15 & 85.41~$\pm$~0.24 & 80.77~$\pm$~0.24\\
    APL& 80.37 & 54.46 & 84.40~$\pm$~0.26 & 84.45~$\pm$~0.50 & 84.35~$\pm$~0.43 & 78.16~$\pm$~0.17\\
    PCE&85.72 & 69.48 & 63.06~$\pm$~0.37 & 62.26~$\pm$~0.36 & 35.47~$\pm$~0.36 & 33.80~$\pm$~0.33\\
    AUL & 82.77 & 66.25 & 76.26 $\pm$ 0.28 & 75.24 $\pm$ 0.20 & 75.48 $\pm$ 0.40 & 63.61 $\pm$ 1.62\\
    CELC & \underline{\textbf{86.38}}  & 69.05 & 89.77~$\pm$~0.39 & 89.19~$\pm$~0.46 & 90.06~$\pm$~0.33 & 81.16~$\pm$~1.86  \\
    Revision& 85.70 & 70.97 & 90.39~$\pm$~0.12 & 90.15~$\pm$~0.11 & 90.07~$\pm$~0.08 & 83.47~$\pm$~0.27 \\
    Identifiability & 82.21 & 67.07 & 82.52 $\pm$ 0.87 & 81.97  $\pm$ 0.85 & 82.09  $\pm$ 0.73 & 71.62  $\pm$ 1.16 \\
    Joint & 84.74 & 70.26 & 88.20~$\pm$~0.29  & 87.54~$\pm$~0.33 & 87.67~$\pm$~0.22 & 84.29~$\pm$~0.40 \\
    Co-teaching & 83.73 & 67.94 & 90.26~$\pm$~0.22 & 89.82~$\pm$~0.63 & 90.64~$\pm$~0.47 & 75.64~$\pm$~4.04   \\
    SIGUA & 79.68 & 65.37 & 87.67~$\pm$~1.18 & 89.01~$\pm$~0.34 & 88.40~$\pm$~0.42 & 80.65~$\pm$~1.29 \\
    Co-Dis & 86.13 & \textbf{71.60} &90.77~$\pm$~0.35 & 90.22~$\pm$~0.30 & 90.35~$\pm$~1.12 & 76.12~$\pm$~3.19\\\midrule
    RT-Catoni's
     & \textbf{86.13} & \underline{\textbf{72.69}} & \textbf{91.31~$\pm$~0.25} & \textbf{91.22~$\pm$~0.40} & \underline{\textbf{91.23~$\pm$~0.41}} & \textbf{84.46~$\pm$~0.41} \\
    RT-Log-sum
     & \textbf{86.15} & \textbf{72.64} & \textbf{91.37~$\pm$~0.07} & \underline{\textbf{91.38~$\pm$~0.23}} & \textbf{91.19~$\pm$~0.10} &  \textbf{85.03~$\pm$~0.54} \\
    RT-Welsch+
     & 85.86 & 70.81 & \underline{\textbf{91.49~$\pm$~0.11}} & \textbf{91.26~$\pm$~0.17} & \textbf{91.09~$\pm$~0.17} & \underline{\textbf{85.96~$\pm$~1.56}}\\
    \bottomrule
    \end{tabular}%
  \caption{\small{Test accuracy (\%) on three real-world noisy datasets, i.e., \textit{Food-101}, \textit{Clothing1M}, and \textit{CIFAR-10N}}. The best 3 experimental results are in bold while the best is underlined.}
  \vspace{-10pt}
  \label{tab:real}%
\end{table*}

\subsubsection{Visualization of experimental results}
We use 2D t-SNE \cite{van2008visualizing} to visualize the experimental results which are presented in Fig.~\ref{fig:ab}. We can see that the proposed methods work well, and can distinguish different classes clearly when there are noisy labels.

\subsubsection{Combination with semi-supervised learning}\label{sec:combine_with_ssl}
Recall that we discussed the comparison between the proposed methods and some methods comprising multiple techniques is unfair. Therefore, to make it fair, here we boost our methods with semi-supervised learning. Specifically, we develop the framework of DivideMix~\cite{li2020dividemix}. Different from the original DivideMix which just uses the cross-entropy loss for follow-up sample selection and semi-supervised learning, we employ our regularly truncated M-estimators for warm-up. Results are provided in Table~\ref{tab:dividemix}. As can be seen, in almost all cases, the proposed methods can bring performance improvements. Especially in the cases of  Pair.-45\%, the improvement is significant.

\subsubsection{Combination with self-supervised learning}\label{sec:combine_with_self}
There are some works that employ self-supervised learning~\cite{wang2022exploring} to enhance network robustness~\cite{li2022selective,wei2021robust}. Hence, here we follow them and show that our methods can be combined with self-supervised learning to enhance network robustness. Specifically, we use MOCO V2~\cite{chen2020improved}. After the self-supervised representation learning, the baseline employs the cross-entropy loss and noisily labeled data to fine-tune the linear head. In contrast, our strategies utilize the regularly truncated M-estimators for fine-tuning. Experimental comparisons are provided in Table~\ref{tab:self}, which demonstrates the utility of our methods. Note that compared with the results in Table~\ref{tab:synthetic}, we claim that one of the advantages of our methods is plug-and-play for robustness improvement.

\subsection{Experiments on real-world noisy datasets}\label{sec:4.3}
\subsubsection{Experimental setup}
\textbf{Datasets.} We exploit three real-world noisy datasets to justify our claims, i.e., \textit{Food-101} \cite{bossard2014food}, \textit{Clothing1M} \cite{xiao2015learning}, and \textit{CIFAR-10N}~\cite{wei2022learning}\footnote{http://competition.noisylabels.com/}, which consist of heterogeneous noisy labels. \textit{Food-101} consists of 101 food categories, with 101,000 images. For each class, 250 manually reviewed clean test images are provided as well as 750 training images.  \textit{Clothing1M} has 1M images with real-world noisy labels, and 50k, 14k, 10k images with clean labels for training, validating, and testing, but with 14 classes. Note that we do not use the 50k and 14k clean data in all the experiments, since it is more practical that there is no available clean data. For preprocessing, we resize the image to 256$\times$256, crop the middle 224$\times$224 as input, and perform normalization. \textit{CIFAR-10N} provides \textit{CIFAR-10} images with human-annotated noisy labels obtained from Amazon Mechanical Turk. Four versions of \textit{CIFAR-10N} label sets are employed here, three of which are labeled by three independent workers (named \textit{CIFAR-10N-1/2/3}) and one of which is negatively aggregated from the above three sets (named \textit{CIFAR-10N-W}). We leave 10\% noisy training data as a validation set for model selection. 

\noindent\textbf{Network structure and optimizer.} We exploit the ResNet-50 network pretrained on ImageNet for \textit{Food-101} and \textit{Clothing1M}. For \textit{Food-101}, we use SGD with momentum 0.9, weight decay $10^{-4}$, batch size 128, and an initial learning rate $10^{-2}$ to train the networks. The learning rate is also divided by 10 after the 40th epoch and 80th epoch. The maximum number of epochs is set to 200. For \textit{Clothing1M}, we also use SGD with momentum 0.9. The batch size and weight decay are adjusted to 32 and $5 \times 10^{-3}$. The learning rate is initially set to $10^{-3}$ and then divided by 10 after the 5th epoch. The maximum number of epochs is set to 20. The experiments on \textit{Food-101} and \textit{Clothing1M} are performed once due to the huge computational cost. For \textit{CIFAR-10N}, a PreAct-ResNet-18 network is exploited. We use
SGD with momentum 0.9, weight decay $10^{-3}$, batch size 128, and an initial learning rate $10^{-2}$. The learning rate is divided by 10 after the 100th epoch. The maximum number of epochs is set to 200. Experiments on \textit{CIFAR-10N} are repeated five times.

\subsubsection{Discussions of experimental results}
Experimental results on real-world noisy datasets are shown in Table \ref{tab:real}. For \textit{Food-101}, the proposed methods achieve great performance. Although the baseline CELC achieves the best performance, the proposed RT-Catoni's and RT-Log-sum achieve competitive performance. For \textit{Clothing1M}, the proposed methods, e.g., RT-Catoni's and RT-Log-sum, achieve clear leads over baselines.  For the proposed RT-Welsch+, although it does not outperform the best baseline Co-Dis, it still receives competitive performance. Moreover, since the training procedure of Co-Dis consists of two stages (estimating the noise rate and performing sample selection), our method can keep an end-to-end manner, and is thus arguably easier to implement. At last, for \textit{CIFAR-10N}, our methods consistently outperform baselines.

\section{Conclusion}\label{sec:5}
In this paper, we focus on exploiting the sample selection approach to handle noisy labels. We discuss that the prior sample selection procedure has some weaknesses, i.e., ignoring the concerns of noisy labels in selected small-loss examples and neglecting the values of discarded large-loss examples. To relieve two issues at the same time, we propose regularly truncated M-estimators, which can assign different weights to selected small-loss examples and enable large-loss examples to periodically participate in optimization. Theoretically, we discuss the noise-tolerant of truncated M-estimators. Empirically, we conduct a series of experiments to verify the effectiveness of the proposed methods. Extensive experimental results support our claims well. In the future, we are interested in applying our method to data cleaning and robustness enhancement of large-scale pre-trained models~\cite{liu2023pre,zhang2023trained}.

{ 
\tiny
\bibliography{bib}

\begin{thebibliography}{100}

\bibitem{zhou2018brief}
Zhi-Hua Zhou.
\newblock A brief introduction to weakly supervised learning.
\newblock {\em National science review}, 5(1):44--53, 2018.

\bibitem{yan2023mutual}
Yan Yan and Yuhong Guo.
\newblock Mutual partial label learning with competitive label noise.
\newblock In {\em ICLR}, 2023.

\bibitem{paleka2023law}
Daniel Paleka and Amartya Sanyal.
\newblock A law of adversarial risk, interpolation, and label noise.
\newblock In {\em ICLR}, 2023.

\bibitem{olmin2022robustness}
Amanda Olmin and Fredrik Lindsten.
\newblock Robustness and reliability when training with noisy labels.
\newblock In {\em AISTATS}, pages 922--942, 2022.

\bibitem{jiang2020beyond}
Lu~Jiang, Di~Huang, Mason Liu, and Weilong Yang.
\newblock Beyond synthetic noise: Deep learning on controlled noisy labels.
\newblock In {\em ICML}, pages 4804--4815, 2020.

\bibitem{he2010maximum}
Ran He, Wei-Shi Zheng, and Bao-Gang Hu.
\newblock Maximum correntropy criterion for robust face recognition.
\newblock {\em IEEE Transactions on Pattern Analysis and Machine Intelligence},
  33(8):1561--1576, 2010.

\bibitem{he2013half}
Ran He, Wei-Shi Zheng, Tieniu Tan, and Zhenan Sun.
\newblock Half-quadratic-based iterative minimization for robust sparse
  representation.
\newblock {\em IEEE Transactions on Pattern Analysis and Machine Intelligence},
  36(2):261--275, 2013.

\bibitem{ke2020laplacian}
Jingchen Ke, Chen Gong, Tongliang Liu, Lin Zhao, Jian Yang, and Dacheng Tao.
\newblock Laplacian welsch regularization for robust semisupervised learning.
\newblock {\em IEEE transactions on cybernetics}, 2020.

\bibitem{patel2023adaptive}
Deep Patel and PS~Sastry.
\newblock Adaptive sample selection for robust learning under label noise.
\newblock In {\em WACV}, pages 3932--3942, 2023.

\bibitem{iscen2022learning}
Ahmet Iscen, Jack Valmadre, Anurag Arnab, and Cordelia Schmid.
\newblock Learning with neighbor consistency for noisy labels.
\newblock In {\em CVPR}, pages 4672--4681, 2022.

\bibitem{bae2022noisy}
HeeSun Bae, Seungjae Shin, Byeonghu Na, JoonHo Jang, Kyungwoo Song, and Il-Chul
  Moon.
\newblock From noisy prediction to true label: Noisy prediction calibration via
  generative model.
\newblock In {\em ICML}, pages 1277--1297, 2022.

\bibitem{liang2022few}
Kevin~J Liang, Samrudhdhi~B Rangrej, Vladan Petrovic, and Tal Hassner.
\newblock Few-shot learning with noisy labels.
\newblock In {\em CVPR}, pages 9089--9098, 2022.

\bibitem{yang2023parametrical}
Shuo Yang, Songhua Wu, Erkun Yang, Bo~Han, Yang Liu, Min Xu, Gang Niu, and
  Tongliang Liu.
\newblock A parametrical model for instance-dependent label noise.
\newblock {\em IEEE Transactions on Pattern Analysis and Machine Intelligence},
  2023.

\bibitem{silva2022noise}
Amila Silva, Ling Luo, Shanika Karunasekera, and Christopher Leckie.
\newblock Noise-robust learning from multiple unsupervised sources of inferred
  labels.
\newblock In {\em AAAI}, volume~36, pages 8315--8323, 2022.

\bibitem{zhang2017understanding}
Chiyuan Zhang, Samy Bengio, Moritz Hardt, Benjamin Recht, and Oriol Vinyals.
\newblock Understanding deep learning requires rethinking generalization.
\newblock In {\em ICLR}, 2017.

\bibitem{li2021provably}
Xuefeng Li, Tongliang Liu, Bo~Han, Gang Niu, and Masashi Sugiyama.
\newblock Provably end-to-end label-noise learning without anchor points.
\newblock In {\em ICML}, 2021.

\bibitem{xia2020part}
Xiaobo Xia, Tongliang Liu, Bo~Han, Nannan Wang, Mingming Gong, Haifeng Liu,
  Gang Niu, Dacheng Tao, and Masashi Sugiyama.
\newblock Part-dependent label noise: Towards instance-dependent label noise.
\newblock In {\em NeurIPS}, 2020.

\bibitem{bai2021me}
Yingbin Bai and Tongliang Liu.
\newblock Me-momentum: Extracting hard confident examples from noisily labeled
  data.
\newblock In {\em ICCV}, 2021.

\bibitem{wu2018light}
Xiang Wu, Ran He, Zhenan Sun, and Tieniu Tan.
\newblock A light cnn for deep face representation with noisy labels.
\newblock {\em IEEE Transactions on Information Forensics and Security},
  13(11):2884--2896, 2018.

\bibitem{xie2022ccmn}
Ming-Kun Xie and Sheng-Jun Huang.
\newblock Ccmn: A general framework for learning with class-conditional
  multi-label noise.
\newblock {\em IEEE Transactions on Pattern Analysis and Machine Intelligence},
  45(1):154--166, 2022.

\bibitem{xie2021partial}
Ming-Kun Xie and Sheng-Jun Huang.
\newblock Partial multi-label learning with noisy label identification.
\newblock {\em IEEE Transactions on Pattern Analysis and Machine Intelligence},
  44(7):3676--3687, 2021.

\bibitem{dai2022towards}
Enyan Dai, Wei Jin, Hui Liu, and Suhang Wang.
\newblock Towards robust graph neural networks for noisy graphs with sparse
  labels.
\newblock In {\em WSDM}, pages 181--191, 2022.

\bibitem{wang2023promix}
Haobo Wang, Ruixuan Xiao, Yiwen Dong, Lei Feng, and Junbo Zhao.
\newblock Promix: combating label noise via maximizing clean sample utility.
\newblock In {\em IJCAI}, 2023.

\bibitem{bucarelli2023leveraging}
Maria~Sofia Bucarelli, Lucas Cassano, Federico Siciliano, Amin Mantrach, and
  Fabrizio Silvestri.
\newblock Leveraging inter-rater agreement for classification in the presence
  of noisy labels.
\newblock In {\em CVPR}, pages 3439--3448, 2023.

\bibitem{yao2020searching}
Quanming Yao, Hansi Yang, Bo~Han, Gang Niu, and James Tin-Yau Kwok.
\newblock Searching to exploit memorization effect in learning with noisy
  labels.
\newblock In {\em ICML}, pages 10789--10798, 2020.

\bibitem{yu2019does}
Xingrui Yu, Bo~Han, Jiangchao Yao, Gang Niu, Ivor~W Tsang, and Masashi
  Sugiyama.
\newblock How does disagreement benefit co-teaching?
\newblock In {\em ICML}, 2019.

\bibitem{feng2023ot}
Chuanwen Feng, Yilong Ren, and Xike Xie.
\newblock Ot-filter: An optimal transport filter for learning with noisy
  labels.
\newblock In {\em CVPR}, 2023.

\bibitem{jiang2018mentornet}
Lu~Jiang, Zhengyuan Zhou, Thomas Leung, Li-Jia Li, and Li~Fei-Fei.
\newblock {MentorNet}: Learning data-driven curriculum for very deep neural
  networks on corrupted labels.
\newblock In {\em ICML}, pages 2309--2318, 2018.

\bibitem{han2018co}
Bo~Han, Quanming Yao, Xingrui Yu, Gang Niu, Miao Xu, Weihua Hu, Ivor Tsang, and
  Masashi Sugiyama.
\newblock Co-teaching: Robust training of deep neural networks with extremely
  noisy labels.
\newblock In {\em NeurIPS}, pages 8527--8537, 2018.

\bibitem{xia2021instance}
Xiaobo Xia, Tongliang Liu, Bo~Han, Mingming Gong, Jun Yu, Gang Niu, and Masashi
  Sugiyama.
\newblock Instance correction for learning with open-set noisy labels.
\newblock {\em arXiv preprint arXiv:2106.00455}, 2021.

\bibitem{wei2020combating}
Hongxin Wei, Lei Feng, Xiangyu Chen, and Bo~An.
\newblock Combating noisy labels by agreement: A joint training method with
  co-regularization.
\newblock In {\em CVPR}, pages 13726--13735, 2020.

\bibitem{arazo2019unsupervised}
Eric Arazo, Diego Ortego, Paul Albert, Noel O’Connor, and Kevin McGuinness.
\newblock Unsupervised label noise modeling and loss correction.
\newblock In {\em ICML}, pages 312--321, 2019.

\bibitem{han2020sigua}
Bo~Han, Gang Niu, Xingrui Yu, Quanming Yao, Miao Xu, Ivor Tsang, and Masashi
  Sugiyama.
\newblock Sigua: Forgetting may make learning with noisy labels more robust.
\newblock In {\em ICML}, 2020.

\bibitem{zhang2014novel}
Teng Zhang and Gilad Lerman.
\newblock A novel m-estimator for robust pca.
\newblock {\em The Journal of Machine Learning Research}, 15(1):749--808, 2014.

\bibitem{zhang2018generalized}
Zhilu Zhang and Mert Sabuncu.
\newblock Generalized cross entropy loss for training deep neural networks with
  noisy labels.
\newblock In {\em NeurIPS}, pages 8778--8788, 2018.

\bibitem{ma2020normalized}
Xingjun Ma, Hanxun Huang, Yisen Wang, Simone Romano, Sarah Erfani, and James
  Bailey.
\newblock Normalized loss functions for deep learning with noisy labels.
\newblock In {\em ICML}, pages 6543--6553, 2020.

\bibitem{lyu2019curriculum}
Yueming Lyu and Ivor~W Tsang.
\newblock Curriculum loss: Robust learning and generalization against label
  corruption.
\newblock In {\em ICLR}, 2020.

\bibitem{wang2019symmetric}
Yisen Wang, Xingjun Ma, Zaiyi Chen, Yuan Luo, Jinfeng Yi, and James Bailey.
\newblock Symmetric cross entropy for robust learning with noisy labels.
\newblock In {\em ICCV}, pages 322--330, 2019.

\bibitem{charoenphakdee2019symmetric}
Nontawat Charoenphakdee, Jongyeong Lee, and Masashi Sugiyama.
\newblock On symmetric losses for learning from corrupted labels.
\newblock In {\em ICML}, pages 961--970, 2019.

\bibitem{kim2019nlnl}
Youngdong Kim, Junho Yim, Juseung Yun, and Junmo Kim.
\newblock Nlnl: Negative learning for noisy labels.
\newblock In {\em ICCV}, pages 101--110, 2019.

\bibitem{liu2019peer}
Yang Liu and Hongyi Guo.
\newblock Peer loss functions: Learning from noisy labels without knowing noise
  rates.
\newblock In {\em ICML}, 2020.

\bibitem{xu2019l_dmi}
Yilun Xu, Peng Cao, Yuqing Kong, and Yizhou Wang.
\newblock L\_dmi: A novel information-theoretic loss function for training deep
  nets robust to label noise.
\newblock In {\em NeurIPS}, pages 6222--6233, 2019.

\bibitem{liu2016classification}
Tongliang Liu and Dacheng Tao.
\newblock Classification with noisy labels by importance reweighting.
\newblock {\em IEEE Transactions on pattern analysis and machine intelligence},
  38(3):447--461, 2016.

\bibitem{ren2018learning}
Mengye Ren, Wenyuan Zeng, Bin Yang, and Raquel Urtasun.
\newblock Learning to reweight examples for robust deep learning.
\newblock In {\em ICML}, pages 4331--4340, 2018.

\bibitem{wu2020class2simi}
Songhua Wu, Xiaobo Xia, Tongliang Liu, Bo~Han, Mingming Gong, Nannan Wang,
  Haifeng Liu, and Gang Niu.
\newblock Class2simi: A noise reduction perspective on learning with noisy
  labels.
\newblock In {\em ICML}, 2021.

\bibitem{xia2022extended}
Xiaobo Xia, Bo~Han, Nannan Wang, Jiankang Deng, Jiatong Li, Yinian Mao, and
  Tongliang Liu.
\newblock Extended {T}: Learning with mixed closed-set and open-set noisy
  labels.
\newblock {\em IEEE Transactions on Pattern Analysis and Machine Intelligence},
  2022.

\bibitem{yao2020dual}
Yu~Yao, Tongliang Liu, Bo~Han, Mingming Gong, Jiankang Deng, Gang Niu, and
  Masashi Sugiyama.
\newblock Dual t: Reducing estimation error for transition matrix in
  label-noise learning.
\newblock In {\em NeurIPS}, 2020.

\bibitem{zhu2021clusterability}
Zhaowei Zhu, Yiwen Song, and Yang Liu.
\newblock Clusterability as an alternative to anchor points when learning with
  noisy labels.
\newblock In {\em ICML}, pages 12912--12923, 2021.

\bibitem{zhu2021second}
Zhaowei Zhu, Tongliang Liu, and Yang Liu.
\newblock A second-order approach to learning with instance-dependent label
  noise.
\newblock In {\em CVPR}, 2021.

\bibitem{liu2022identifiability}
Yang Liu.
\newblock Identifiability of label noise transition matrix.
\newblock {\em arXiv preprint arXiv:2202.02016}, 2022.

\bibitem{kye2022learning}
Seong~Min Kye, Kwanghee Choi, Joonyoung Yi, and Buru Chang.
\newblock Learning with noisy labels by efficient transition matrix estimation
  to combat label miscorrection.
\newblock In {\em ECCV}, pages 717--738, 2022.

\bibitem{goldberger2016training}
Jacob Goldberger and Ehud Ben-Reuven.
\newblock Training deep neural-networks using a noise adaptation layer.
\newblock In {\em ICLR}, 2017.

\bibitem{yi2019probabilistic}
Kun Yi and Jianxin Wu.
\newblock Probabilistic end-to-end noise correction for learning with noisy
  labels.
\newblock In {\em CVPR}, pages 7017--7025, 2019.

\bibitem{zhang2021learningwith}
Yikai Zhang, Songzhu Zheng, Pengxiang Wu, Mayank Goswami, and Chao Chen.
\newblock Learning with feature-dependent label noise: A progressive approach.
\newblock In {\em ICLR}, 2021.

\bibitem{xiao2015learning}
Tong Xiao, Tian Xia, Yi~Yang, Chang Huang, and Xiaogang Wang.
\newblock Learning from massive noisy labeled data for image classification.
\newblock In {\em CVPR}, pages 2691--2699, 2015.

\bibitem{vahdat2017toward}
Arash Vahdat.
\newblock Toward robustness against label noise in training deep discriminative
  neural networks.
\newblock In {\em NeurIPS}, 2017.

\bibitem{li2017learning}
Yuncheng Li, Jianchao Yang, Yale Song, Liangliang Cao, Jiebo Luo, and Li-Jia
  Li.
\newblock Learning from noisy labels with distillation.
\newblock In {\em ICCV}, pages 1910--1918, 2017.

\bibitem{tanaka2018joint}
Daiki Tanaka, Daiki Ikami, Toshihiko Yamasaki, and Kiyoharu Aizawa.
\newblock Joint optimization framework for learning with noisy labels.
\newblock In {\em CVPR}, 2018.

\bibitem{li2020dividemix}
Junnan Li, Richard Socher, and Steven~C.H. Hoi.
\newblock Dividemix: Learning with noisy labels as semi-supervised learning.
\newblock In {\em ICLR}, 2020.

\bibitem{li2022selective}
Shikun Li, Xiaobo Xia, Shiming Ge, and Tongliang Liu.
\newblock Selective-supervised contrastive learning with noisy labels.
\newblock In {\em CVPR}, pages 316--325, 2022.

\bibitem{huang2023twin}
Zhizhong Huang, Junping Zhang, and Hongming Shan.
\newblock Twin contrastive learning with noisy labels.
\newblock In {\em CVPR}, pages 11661--11670, 2023.

\bibitem{zhang2017mixup}
Hongyi Zhang, Moustapha Cisse, Yann~N Dauphin, and David Lopez-Paz.
\newblock mixup: Beyond empirical risk minimization.
\newblock {\em arXiv preprint arXiv:1710.09412}, 2017.

\bibitem{reed2014training}
Scott~E Reed, Honglak Lee, Dragomir Anguelov, Christian Szegedy, Dumitru Erhan,
  and Andrew Rabinovich.
\newblock Training deep neural networks on noisy labels with bootstrapping.
\newblock In {\em ICLR}, 2015.

\bibitem{berthelot2019mixmatch}
David Berthelot, Nicholas Carlini, Ian Goodfellow, Nicolas Papernot, Avital
  Oliver, and Colin Raffel.
\newblock Mixmatch: A holistic approach to semi-supervised learning.
\newblock {\em arXiv preprint arXiv:1905.02249}, 2019.

\bibitem{wang2023mosaic}
Zhaoqing Wang, Ziyu Chen, Yaqian Li, Yandong Guo, Jun Yu, Mingming Gong, and
  Tongliang Liu.
\newblock Mosaic representation learning for self-supervised visual
  pre-training.
\newblock In {\em ICLR}, 2023.

\bibitem{han2020survey}
Bo~Han, Quanming Yao, Tongliang Liu, Gang Niu, Ivor~W Tsang, James~T Kwok, and
  Masashi Sugiyama.
\newblock A survey of label-noise representation learning: Past, present and
  future.
\newblock {\em arXiv preprint arXiv:2011.04406}, 2020.

\bibitem{song2022learning}
Hwanjun Song, Minseok Kim, Dongmin Park, Yooju Shin, and Jae-Gil Lee.
\newblock Learning from noisy labels with deep neural networks: A survey.
\newblock {\em IEEE Transactions on Neural Networks and Learning Systems},
  2022.

\bibitem{patrini2017making}
Giorgio Patrini, Alessandro Rozza, Aditya Krishna~Menon, Richard Nock, and
  Lizhen Qu.
\newblock Making deep neural networks robust to label noise: A loss correction
  approach.
\newblock In {\em CVPR}, 2017.

\bibitem{mohri2018foundations}
Mehryar Mohri, Afshin Rostamizadeh, and Ameet Talwalkar.
\newblock {\em Foundations of Machine Learning}.
\newblock MIT Press, 2018.

\bibitem{xia2021sample}
Xiaobo Xia, Tongliang Liu, Bo~Han, Mingming Gong, Jun Yu, Gang Niu, and Masashi
  Sugiyama.
\newblock Sample selection with uncertainty of losses for learning with noisy
  labels.
\newblock In {\em ICLR}, 2022.

\bibitem{wang2018iterative}
Yisen Wang, Weiyang Liu, Xingjun Ma, James Bailey, Hongyuan Zha, Le~Song, and
  Shu-Tao Xia.
\newblock Iterative learning with open-set noisy labels.
\newblock In {\em CVPR}, pages 8688--8696, 2018.

\bibitem{cheng2017learning}
Jiacheng Cheng, Tongliang Liu, Kotagiri Ramamohanarao, and Dacheng Tao.
\newblock Learning with bounded instance-and label-dependent label noise.
\newblock In {\em ICML}, 2020.

\bibitem{james2013introduction}
Gareth James, Daniela Witten, Trevor Hastie, and Robert Tibshirani.
\newblock {\em An introduction to statistical learning}, volume 112.
\newblock Springer, 2013.

\bibitem{he2013robust}
Ran He, Tieniu Tan, and Liang Wang.
\newblock Robust recovery of corrupted low-rankmatrix by implicit regularizers.
\newblock {\em IEEE Transactions on Pattern Analysis and Machine Intelligence},
  36(4):770--783, 2013.

\bibitem{catoni2012challenging}
Olivier Catoni.
\newblock Challenging the empirical mean and empirical variance: a deviation
  study.
\newblock In {\em Annales de l'IHP Probabilit{\'e}s et statistiques},
  volume~48, pages 1148--1185, 2012.

\bibitem{candes2008enhancing}
Emmanuel~J Candes, Michael~B Wakin, and Stephen~P Boyd.
\newblock Enhancing sparsity by reweighted l1 minimization.
\newblock {\em Journal of Fourier analysis and applications}, 14(5-6):877--905,
  2008.

\bibitem{liu2007correntropy}
Weifeng Liu, Puskal~P Pokharel, and Jose~C Principe.
\newblock Correntropy: Properties and applications in non-gaussian signal
  processing.
\newblock {\em IEEE Transactions on signal processing}, 55(11):5286--5298,
  2007.

\bibitem{ghosh2017robust}
Aritra Ghosh, Himanshu Kumar, and P~Shanti Sastry.
\newblock Robust loss functions under label noise for deep neural networks.
\newblock In {\em AAAI}, 2017.

\bibitem{durrett2019probability}
Rick Durrett.
\newblock {\em Probability: theory and examples}, volume~49.
\newblock Cambridge university press, 2019.

\bibitem{guan2017truncated}
Naiyang Guan, Tongliang Liu, Yangmuzi Zhang, Dacheng Tao, and Larry~S Davis.
\newblock Truncated cauchy non-negative matrix factorization.
\newblock {\em IEEE Transactions on Pattern Analysis and Machine Intelligence},
  41(1):246--259, 2017.

\bibitem{nagy2006parameter}
Ferenc Nagy.
\newblock Parameter estimation of the cauchy distribution in information theory
  approach.
\newblock {\em J. UCS}, 12(9):1332--1344, 2006.

\bibitem{menon2020can}
Aditya~Krishna Menon, Ankit~Singh Rawat, Sashank~J. Reddi, and Sanjiv Kumar.
\newblock Can gradient clipping mitigate label noise?
\newblock In {\em ICLR}, 2020.

\bibitem{zhou2023asymmetric}
Xiong Zhou, Xianming Liu, Deming Zhai, Junjun Jiang, and Xiangyang Ji.
\newblock Asymmetric loss functions for noise-tolerant learning: Theory and
  applications.
\newblock {\em IEEE Transactions on Pattern Analysis and Machine Intelligence},
  2023.

\bibitem{wei2023mitigating}
Hongxin Wei, Huiping Zhuang, Renchunzi Xie, Lei Feng, Gang Niu, Bo~An, and
  Yixuan Li.
\newblock Mitigating memorization of noisy labels by clipping the model
  prediction.
\newblock In {\em ICML}, 2023.

\bibitem{xia2019anchor}
Xiaobo Xia, Tongliang Liu, Nannan Wang, Bo~Han, Chen Gong, Gang Niu, and
  Masashi Sugiyama.
\newblock Are anchor points really indispensable in label-noise learning?
\newblock In {\em NeurIPS}, 2019.

\bibitem{liu2023identifiability}
Yang Liu, Hao Cheng, and Kun Zhang.
\newblock Identifiability of label noise transition matrix.
\newblock In {\em ICML}, pages 21475--21496, 2023.

\bibitem{xia2023co}
Xiaobo Xia, Bo~Han, Yibing Zhan, Jun Yu, Mingming Gong, Chen Gong, and
  Tongliang Liu.
\newblock Combating noisy labels with sample selection by mining
  high-discrepancy examples.
\newblock In {\em ICCV}, 2023.

\bibitem{nguyen2020self}
Duc~Tam Nguyen, Chaithanya~Kumar Mummadi, Thi Phuong~Nhung Ngo, Thi Hoai~Phuong
  Nguyen, Laura Beggel, and Thomas Brox.
\newblock Self: Learning to filter noisy labels with self-ensembling.
\newblock In {\em ICLR}, 2020.

\bibitem{LeCunmnist}
Yann LeCun, Corinna Cortes, and Christopher~J.C. Burges.
\newblock The {MNIST} database of handwritten digits.

\bibitem{netzer2011svhn}
Yuval Netzer, Tao Wang, Adam Coates, Alessandro Bissacco, Bo~Wu, and Andrew
  Y.Ng.
\newblock Reading digits in natural images with unsupervised feature learning.
\newblock In {\em NIPS Workshop on Deep Learning and Unsupervised Feature
  Learning}, 2011.

\bibitem{krizhevsky2009learning}
Alex Krizhevsky.
\newblock Learning multiple layers of features from tiny images.
\newblock Technical report, 2009.

\bibitem{lang1995newsweeder}
Ken Lang.
\newblock Newsweeder: Learning to filter netnews.
\newblock In {\em Machine Learning Proceedings}, pages 331--339. 1995.

\bibitem{lee2019robust}
Kimin Lee, Sukmin Yun, Kibok Lee, Honglak Lee, Bo~Li, and Jinwoo Shin.
\newblock Robust inference via generative classifiers for handling noisy
  labels.
\newblock In {\em ICML}, pages 3763--3772, 2019.

\bibitem{pennington2014glove}
Jeffrey Pennington, Richard Socher, and Christopher~D Manning.
\newblock Glove: Global vectors for word representation.
\newblock In {\em EMNLP}, 2014.

\bibitem{xia2021robust}
Xiaobo Xia, Tongliang Liu, Bo~Han, Chen Gong, Nannan Wang, Zongyuan Ge, and
  Yi~Chang.
\newblock Robust early-learning: Hindering the memorization of noisy labels.
\newblock In {\em ICLR}, 2021.

\bibitem{arpit2017closer}
Devansh Arpit, Stanis{\l}aw Jastrz{\k{e}}bski, Nicolas Ballas, David Krueger,
  Emmanuel Bengio, Maxinder~S Kanwal, Tegan Maharaj, Asja Fischer, Aaron
  Courville, Yoshua Bengio, et~al.
\newblock A closer look at memorization in deep networks.
\newblock In {\em ICML}, pages 233--242, 2017.

\bibitem{van2008visualizing}
Laurens Van~der Maaten and Geoffrey Hinton.
\newblock Visualizing data using t-sne.
\newblock {\em Journal of machine learning research}, 9(11), 2008.

\bibitem{wang2022exploring}
Zhaoqing Wang, Qiang Li, Guoxin Zhang, Pengfei Wan, Wen Zheng, Nannan Wang,
  Mingming Gong, and Tongliang Liu.
\newblock Exploring set similarity for dense self-supervised representation
  learning.
\newblock In {\em CVPR}, pages 16590--16599, 2022.

\bibitem{wei2021robust}
Tong Wei, Jiang-Xin Shi, Wei-Wei Tu, and Yu-Feng Li.
\newblock Robust long-tailed learning under label noise.
\newblock {\em arXiv preprint arXiv:2108.11569}, 2021.

\bibitem{chen2020improved}
Xinlei Chen, Haoqi Fan, Ross Girshick, and Kaiming He.
\newblock Improved baselines with momentum contrastive learning.
\newblock {\em arXiv preprint arXiv:2003.04297}, 2020.

\bibitem{bossard2014food}
Lukas Bossard, Matthieu Guillaumin, and Luc Van~Gool.
\newblock Food-101--mining discriminative components with random forests.
\newblock In {\em ECCV}, pages 446--461, 2014.

\bibitem{wei2022learning}
Jiaheng Wei, Zhaowei Zhu, Hao Cheng, Tongliang Liu, Gang Niu, and Yang Liu.
\newblock Learning with noisy labels revisited: A study using real-world human
  annotations.
\newblock In {\em ICLR}, 2022.

\bibitem{liu2023pre}
Pengfei Liu, Weizhe Yuan, Jinlan Fu, Zhengbao Jiang, Hiroaki Hayashi, and
  Graham Neubig.
\newblock Pre-train, prompt, and predict: A systematic survey of prompting
  methods in natural language processing.
\newblock {\em ACM Computing Surveys}, 55(9):1--35, 2023.

\bibitem{zhang2023trained}
Ruiqi Zhang, Spencer Frei, and Peter~L Bartlett.
\newblock Trained transformers learn linear models in-context.
\newblock {\em arXiv preprint arXiv:2306.09927}, 2023.

\end{thebibliography}
}

\newpage
\onecolumn
\appendices
\section{Supplementary Theoretical Analysis}
\subsection{Preliminary knowledge}
We denote the underlying clean dataset corresponding to the noisy dataset $S$, as $S^*=\{(\bm{x}_i,y_i)\}_{i=1}^n$, where $y_i$ is the unobserved clean label of $\tilde{y}_i$. Given any loss function, $\psi$, and a classifier, $f$, we define the $\psi$-risk 
\begin{equation}
    R_\psi(f):=\mathbbm{E}_{(\bm{x},y)\sim S^*}[\psi(f(\bm{x}),y)].
\end{equation}
Under the risk minimization framework, the
objective is to learn a classifier, $f$, which is a global minimizer of $R_\psi$. Note that
the $\psi$-risk, $R_\psi$, depends on $\psi$, the loss function. When $\psi$ happens to be the 0–1
loss, $R_\psi$ would be the usual Bayes risk. Let $f^*$ be the
global minimizer (over the chosen function class) of $R_\psi(f)$. In this paper, $\psi$ will be the loss function composed of the cross-entropy loss and truncated M-estimators. 

Then the notion and notations about the label noise model are introduced. We have 
\begin{equation}
    \tilde{y}_i=\begin{cases}
    y_i & \text{with probability}~(1-\eta_{\bm{x}_i})\\
    j, j\in[k], j\neq y_i& \text{with probability}~\bar{\eta}_{\bm{x},j}.
    \end{cases}
\end{equation}
Note that for all $\bm{x}$, conditioned on $y=i$, we have $\sum_{j\neq i}\bar{\eta}_{\bm{x},j}=\eta_{\bm{x}}$. The label noise model is termed \textit{symmetric} or \textit{uniform} if $\eta_{\bm{x}}=\eta$, and $\bar{\eta}_{\bm{x},j}=\frac{\eta}{k-1}$, $\forall j\neq y$, $\forall \bm{x}$, where $\eta$ is a constant. Noise is said to be \textit{simple} non-uniform when the noise rate $\eta_{\bm{x}}$ is a function of $\bm{x}$. A simple special case is when $\bar{\eta}_{\bm{x},j}=\frac{\eta_{\bm{x}}}{k-1}$, $\forall j\neq y$. We 
define it as simple non-uniform noise. Then $\psi$-risk of a classifier
$f$ under noisy data is defined 
\begin{equation}
    R_\psi^\eta(f):=\mathbbm{E}_{(\bm{x},\tilde{y})\sim S}[\psi(f(\bm{x}),\tilde{y})].
\end{equation}
\subsection{Proof of Lemma~1}
Our proofs are inspired by~\cite{ghosh2017robust}. Recall that $R_\psi(f):=\mathbbm{E}_{(\bm{x},y)\sim S^*}[\psi(f(\bm{x}),y)]$. For symmetric noise, we have, for any $f$, 
\begin{align}
    R_\psi^\eta(f)&=\mathbbm{E}_{(\bm{x},\tilde{y})\sim S}[\psi(f(\bm{x}),\tilde{y})]\\\nonumber
    &=\mathbbm{E}_{\bm{x}} \mathbbm{E}_{y|\bm{x}}\mathbbm{E}_{\tilde{y}|\bm{x},y}\psi(f(\bm{x}),\tilde{y})\\\nonumber
    &=\mathbbm{E}_{\bm{x}} \mathbbm{E}_{y|\bm{x}}[(1-\eta)\psi(f(\bm{x}),y)+\frac{\eta}{k-1}\sum_{i\neq y}\psi(f(\bm{x}),i)].
\end{align}
Note that in this paper, $\psi$ will be the loss function that is composed of the cross-entropy loss and truncated M-estimators. Although the cross-entropy loss is not upper-bounded, with our truncation mechanism, $\psi$ will be upper-bounded, since the largest value of $\psi$ is limited. Therefore, for any $\psi$, we denote its lower and upper bounds of the sum of loss values as $c_1$ and $c_2$ respectively, i.e., $c_1\leq\sum_{i}\psi(f(\bm{x}),i)\leq c_2$. In this way, 
\begin{align}
    &\quad R_\psi^\eta(f^*)-R_\psi^\eta(f)\\\nonumber
    &\leq \frac{c_2\eta}{k-1} + (1-\frac{\eta k}{k-1})R_\psi(f^*)-\frac{c_1\eta}{k-1}-(1-\frac{\eta k}{k-1})R_\psi(f)\\\nonumber
    &\leq\frac{(c_2-c_1)\eta}{k-1} + \frac{k-1-\eta k}{k-1}(R_\psi(f^*)-R_\psi(f))\\\nonumber
    &=\frac{(c_2-c_1)\eta}{k-1} + \frac{k-1-\eta k}{k-1}\Delta(\psi,f)\\\nonumber
    &=\frac{(c_2-c_1-k\Delta(\psi,f))\eta+(k-1)\Delta(\psi,f)}{k-1}
\end{align}
If the noise rate $\eta<\frac{(1-k)\Delta(\psi,f)}{c_2-c_1-k\Delta(\psi,f)}$, we will have 
\begin{align}
    &\quad R_\psi^\eta(f^*)-R_\psi^\eta(f)\\\nonumber
    &\leq\frac{(1-k)\Delta(\psi,f)+(k-1)\Delta(\psi,f)}{k-1}\\\nonumber
    &=0.
\end{align}
This proves $f*$ is also a minimizer of the risk under symmetric noise. Proof completed. 

\subsection{Proof of Corollary 1}
\begin{corollary}
    In a multi-class classification problem, the truncated M-estimators are noise-tolerant under the simple non-uniform noise, if $c_2-c_1-k\Delta(\psi,f,\bm{x})>0$ and the noise rate $\eta_{\bm{x}}<\frac{(1-k)\Delta(\psi,f,\bm{x})}{c_2-c_1-k\Delta(\psi,f,\bm{x})}$. Here $c_1$ and $c_2$ denote the lower and upper bounds of the sum of the losses obtained by predictions on all classes, and $\Delta(\psi,f,\bm{x})=\sup (\psi(f^*(\bm{x}), y) - \psi(f(\bm{x}), y))$.
\end{corollary}

The proof of Corollary 1 is as follows. For the simple non-uniform noise, we derive that
\begin{align}
    R_\psi^\eta(f)&=\mathbbm{E}_{\bm{x}} \mathbbm{E}_{y|\bm{x}}[(1-\eta_{\bm{x}}) \psi(f(\bm{x}),y)+\frac{\eta_{\bm{x}}}{k-1}\sum_{i\neq y}\psi(f(\bm{x}),i)]\\\nonumber
    &=\mathbbm{E}_{\bm{x}} \mathbbm{E}_{y|\bm{x}}[(1-\eta_{\bm{x}}) \psi(f(\bm{x}),y)\\\nonumber
    &+\frac{\eta_{\bm{x}}}{k-1}(\sum_i\psi(f(\bm{x}),i)-\psi(f(\bm{x}),y)]\\\nonumber
    &=\mathbbm{E}[\frac{k-1-\eta_{\bm{x}}k}{k-1}\psi(f(\bm{x}),y)+\frac{\eta_{\bm{x}}}{k-1}\sum_i\psi(f(\bm{x}),i)].
\end{align}
Therefore, 
\begin{align}
    &\quad R_\psi^\eta(f^*)-R_\psi^\eta(f)\\\nonumber
    &\leq \frac{(c_2-c_1-k\Delta(\psi,f,\bm{x})\eta+(k-1)\Delta(\psi,f,\bm{x})}{k-1},
\end{align}
where $\Delta(\psi,f,\bm{x})=\sup (\psi(f^*(\bm{x}), y) - \psi(f(\bm{x}), y))$. Therefore, if the noise rate $\eta_{\bm{x}}<\frac{(1-k)\Delta(\psi,f,\bm{x})}{c_2-c_1-k\Delta(\psi,f,\bm{x})}$, we have $R_\psi^\eta(f^*)-R_\psi^\eta(f)\leq 0$. This proves $f^*$ is also a minimizer of the risk under
the simple non-uniform noise.

\end{document}